\documentclass[final]{elsarticle}

\usepackage[utf8]{inputenc} 
\usepackage{lineno}

\usepackage[labelfont=bf,singlelinecheck=false]{caption}
\usepackage[hidelinks]{hyperref}       
\usepackage{url}            

\usepackage{graphicx}
\usepackage{rotating}
\usepackage{color}

\usepackage{fancyhdr}
\begin{document}

\begin{frontmatter}
\title{Grape detection, segmentation, and tracking using \\ deep neural
  networks and three-dimensional association}

\author[1]{Thiago T.~Santos\corref{cor1}}
\ead{thiago.santos@embrapa.br}

\author[2]{Leonardo L.~de~Souza}
\ead{leonardolimasza@gmail.com}

\author[2]{Andreza A. dos Santos}
\ead{andi.apsantos@gmail.com}

\author[2]{Sandra Avila}
\ead{sandra@ic.unicamp.br}

\cortext[cor1]{Corresponding author}
  
\address[1]{
  Embrapa Agricultural Informatics,
  Av. André Tosello 209,
  Campinas SP, 13083-886, Brazil
}

\address[2]{
  Institute of Computing, University of Campinas,
  Av. Albert Einstein 1251,
  Campinas SP, 13083-852, Brazil
}

\begin{abstract}
Agricultural applications such as yield prediction, precision agriculture
and automated harvesting need systems able to infer the crop state
from low-cost sensing devices. Proximal sensing using affordable
cameras combined with computer vision has seen a promising
alternative, strengthened after the advent of convolutional neural
networks (CNNs) as an alternative for challenging pattern recognition
problems in natural images. Considering fruit growing monitoring
and automation, a fundamental problem is the detection, segmentation
and counting of individual fruits in orchards. Here we show that for
wine grapes, a crop presenting large variability in shape, color, size and
compactness, grape clusters can be successfully detected, segmented and
tracked using state-of-the-art CNNs. In {a test set} containing 408 grape
clusters from images taken on {a trellis-system based vineyard}, we have
reached an $F_1$-score up to 0.91 for instance segmentation, a fine
separation of each cluster from other structures in the image  that
allows a more accurate assessment of fruit size and shape. We have
also shown as clusters can be identified and tracked along video sequences recording orchard
rows. We also present a public dataset containing grape clusters
properly annotated in 300 images and a novel annotation methodology for 
segmentation of complex objects in
 natural images. The presented pipeline  for annotation,
training, evaluation and tracking of agricultural patterns in images
can be replicated for different crops and production systems. It
can be employed in the development of sensing components for several
agricultural and environmental applications. 
\end{abstract}
\end{frontmatter}

\pagestyle{fancy}


\section{Introduction}

Automation in agriculture is particularly hard when compared to
industrial automation due to field conditions and the uncertainty
regarding plant structure and outdoor environment. That creates a need   
for systems able to monitor structures as plants and fruits in a
fine-grained level \citep{Kirkpatrick2019}. Proper detection and
localization for such structures are critical components for monitoring, 
robotics and autonomous systems for agriculture \citep{UKRAS2018}. 

Accurate fruit detection and localization are essential
for several applications. Fruit counting and yield estimation are the
more immediate ones. Precision agriculture applications, accounting
for management of inter and intra-field variability, can be derived if
detection data is properly localized in space. Fruit detection can also be a
preliminary step for disease and nutrient deficiency monitoring
\citep{Barbedo2019} and a crucial component on actuation, for example, automated
spraying and harvesting may be an important application considering the
declining in agricultural labor force \citep{Roser}. Beyond farms,
fruit detection can be employed in field phenotyping, aiding plant
research and breeding programs \citep{ Kicherer2017, Rose2016}.

Off-the-shelf RGB cameras and computer vision can provide
affordable and versatile solutions for fruit
detection. State-of-the-art computer vision systems based on deep
convolutional neural networks \citep{LeCun2015} can deal with
variations in pose, shape, illumination and large inter-class variability
\citep{He2016, NIPS2012_4824, Simonyan2014}, essential features needed for robust
recognition of complex objects in outdoor environments. Recent researches
\citep{Bargoti2017b, Sa2016} have shown that the Faster R-CNN
(region-based convolutional neural network) architecture
\citep{NIPS2015_5638} can produce accurate 
results for a large set of fruits, including peppers, melons, oranges,
apples, mangoes, avocados, strawberries and almonds. Detection results
can be integrated by data association approaches, by employing object
tracking or mapping, to perform fruit counting for rows in the crop
field \citep{Liu2019}.

\begin{figure}
  \centering
  \begin{tabular}{cc}
    Input & Semantic Segmentation\\
    \includegraphics[width=0.45\textwidth]{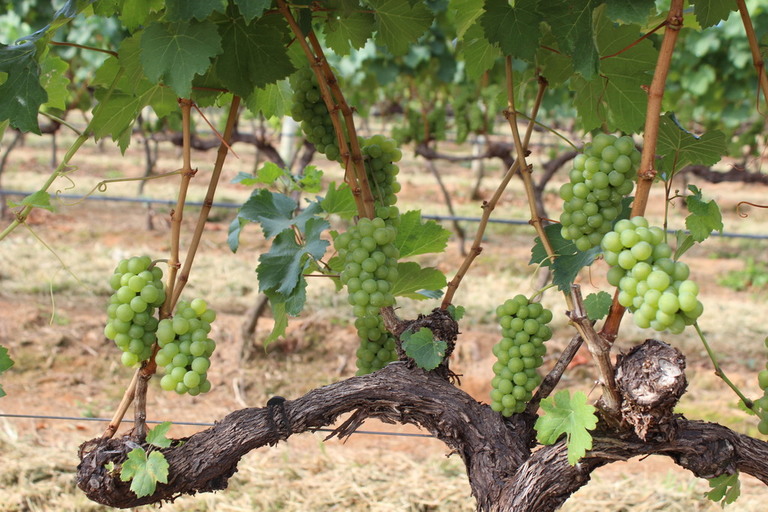} & \includegraphics[width=0.45\textwidth]{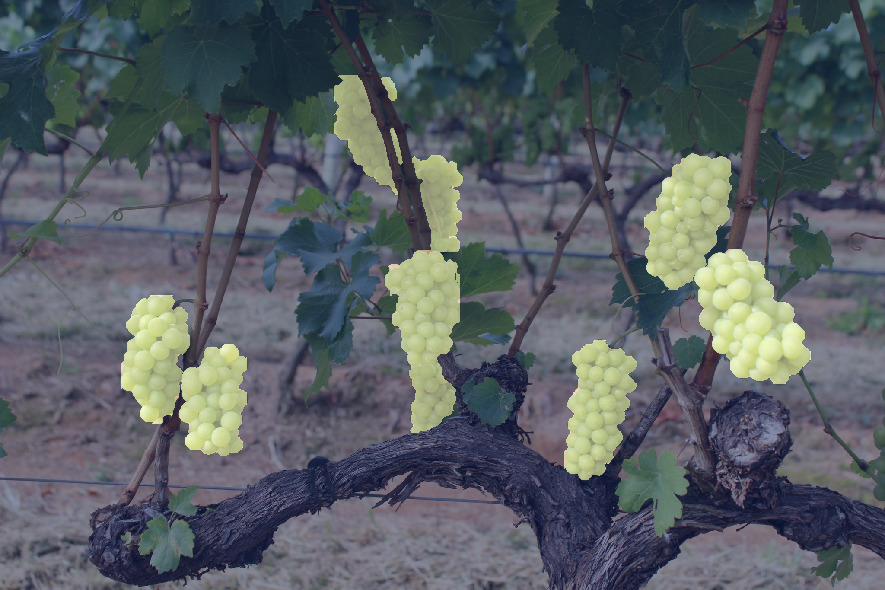}\\ 
    Object Detection & Instance Segmentation\\
    \includegraphics[width=0.45\textwidth]{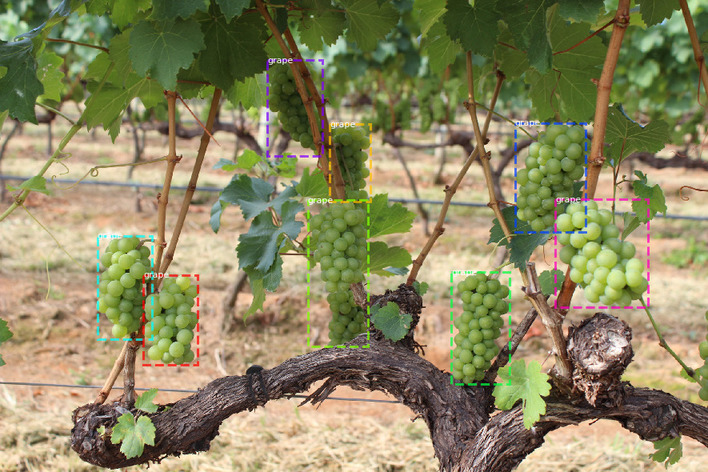} & \includegraphics[width=0.45\textwidth]{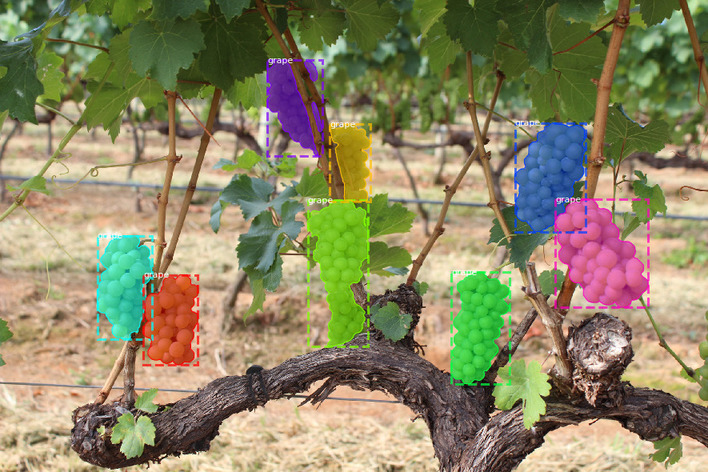}\\ 
  \end{tabular}
  \caption{Dataset entry example. Visual recognition can be stated as
    three different problems: (i) semantic segmentation (a pixel classification problem
    for fruit/non-fruit), (ii) object detection (fruit localization by
    bounding boxes) and (iii) instance segmentation. The most
    challenging variation, instance segmentation, is object detection
    and pixel attribution combined (each pixel is attributed to one of the
    detected objects or to the background) \citep{MSCOCO}.}
  \label{fig:dataset-example}
\end{figure}

These previous detection systems identify individual objects by rectangular
bounding boxes, as seen in Figure~\ref{fig:dataset-example}. Such
boxes, if well fitted to the fruits boundaries, could
provide estimations of fruit shape and space occupancy for fruits
presenting a regular shape as oranges and apples (circular
shape). However, for grape clusters, rectangular boxes would not
properly adjust to the berries. A step further beyond object detection
is \emph{instance segmentation} \citep{MSCOCO}: the fruit/non-fruit pixel
classification combined with instance assignment
(Figure~\ref{fig:dataset-example}). Instance segmentation 
can properly identify berries pixels in the detection box,
providing finer fruit characterization. Also, \emph{occlusions} by leaves,
branches, trunks and even other clusters can be properly addressed by
instance segmentation, aiding on robotic manipulation and other
automation tasks.

To approach fruit instance segmentation as a supervised machine
learning problem, we need datasets that capture the variations
observed in the field. Wine grapes present large variations in shape,
size, color and structure, even for the same grape variety,
contrasting to citrus and apples. Also, the dataset have to provide
\emph{masks} for the individual clusters, isolating grapes from
background pixels and from occluding objects. We also need a neural
network architecture able to simultaneously perform object detection
and pixel classification. Thus, the present work introduces the
following contributions:  

\begin{enumerate}
  \item a new methodology for image annotation that employs interactive
    image segmentation \citep{Noma2012} to generate object masks,
    identifying background and occluding foreground pixels;
  \item a new public dataset\footnote{{Available at
      \url{doi:10.5281/zenodo.3361736} \citep{ZenodoWGISD}.}}
    for grape detection and instance segmentation, 
    comprising images, bounding boxes and masks -- this dataset is
    composed by images of five different grape varieties taken on
    field (Figure~\ref{fig:dataset-example}); 
  \item an evaluation of two deep learning detection architectures for
    grape detection:  Mask~R-CNN \citep{He_2017_ICCV}, a
    convolutional framework for instance segmentation that is simple
    to train and generalizes well \citep{Liu2018}, and YOLO
  \citep{Redmon2016}, a single-stage network 
    that can detect objects without a previous region-proposal stage 
    \citep{Huang2017} -- such evaluation allows a comparison between
    instance segmentation and box-based object detection approaches;
  \item a fruit counting methodology that employs three-dimensional
    association to integrate and localize the detection results in
    space, avoiding multiple counting, addressing occlusions and
    accumulating evidence from different images to confirm detections.  
\end{enumerate}

\section{Related Work}
\label{sec:RelWork}

As seen in computer vision applications on other fields, classic machine learning
and pattern recognition have been replaced by modern deep learning techniques, which can
address the enormous variability in object appearance, as shortly described
in the following sections.

\subsection{Earlier works: feature engineering and machine learning}

Earlier works in fruit detection employed the classic \emph{feature
  engineering} approach: human-designed descriptors based on color,
geometric and texture features. Using such features, machine learning
techniques such as Bayesian classifiers, support vector machines and
clustering were applied to perform fruit detection and
classification. \citet{Gongal2015} presented an
extensive review of the works employing this approach. \citet{Dunn2004} presented one of the
earliest works to employ image processing for grape detection. They used color
thresholding to detect ``grape pixels'' in images showing mature {\it
  Cabernet Sauvignon} clusters in vines. A white screen was placed
behind the canopy to create a regular background.

\citet{Nuske2011} presented a computer vision methodology
intended for realistic field operation without background
control. Their multi-stage method employed Loy and Zelinsky's Radial
Symmetry Transform \citep{Loy2003} to find berry candidates, further filtered
by a K-nearest neighbors classifier using color and texture
features. In the last step, neighboring berries were grouped in
clusters, eliminating isolated berries (likely
false-positives). For a set of 2,973 berries, their system reached 
63.7\% for recall and 98.0\% for precision overall (the set included
berries of three grape varieties). The authors performed linear regression
for the berries count found for individual vines, finding a $0.74$ correlation score
for crop weight.

In a further work, \citet{Nuske2014} added a data
association component based on visual odometry \citep{Scaramuzza2011}
to avoid double-counting and to estimate the spatial distribution of
yield. They proposed a new berry detector for a particular flash-based
setting for night imaging developed by them and evaluated other
image features for berry classification: SIFT \citep{LoweSIFT2004} and
FREAK \citep{Alahi2012}. \citet{Nuske2014} stated that segmentation of berries
clusters (grape clusters) is challenging because of occlusion and touching clusters; after some experiments with 3-D modeling, the authors
chose to perform yield estimation using berry counting. 
They performed controlled imaging and reported variations in results
possibly caused by illumination and imaging differences.

\subsection{Deep learning-based works}

Earlier works present issues that foreshadow the advantages and
power of convolutional neural networks (CNNs). These networks learn
effective representations for a given machine learning task, replacing
feature engineering \citep{Bengio2013}. Systematically, deep learning
approaches are being 
adopted in fields presenting image-based perceptual problems, and
agricultural applications are no exception \citep{Kamilaris2018}.   

CNN's \emph{invariance to local translation} give vision systems
robustness in situations where a feature's presence is 
more important than its exact location
\citep{Goodfellow-et-al-2016}. As an example, 
\citet{Nuske2014} reported that variations in the berry candidate
location by detection affected their berry 
classification. CNNs are also able to encode variance regarding pose,
color and illumination, if the training data presents sufficient
examples of such variation, which relieves the need for controlled
imaging, illumination and camera settings. The first attempts employed 
CNNs to perform pixel classification, followed by additional steps to segment
individual fruits \citep{Bargoti2017a, Chen2017}. Further, these earlier
approaches were replaced by \emph{end-to-end object detection}
\citep{Bargoti2017b, Liu2019, Sa2016}  based on the popular Faster R-CNN
architecture \citep{NIPS2015_5638}.

\citet{Sa2016} employed transfer learning, using a
VGG16 network \citep{Simonyan2014} pre-trained using  ImageNet \citep{Deng2009ImageNet} 
(VGG16 is the \emph{perceptual backbone} in the Faster R-CNN architecture). They
reached $F_1$-scores up to 0.83  in tests on sweet pepper and rock melon,
using a dataset of images captured in a greenhouse, and presented similar
performance for smaller datasets of strawberry, apple, avocado, mango
and orange images retrieved from Google Images Search. The authors also
fused RGB and Near Infrared (NIR) data in four-channel arrays, showing that the CNN
paradigm can easily benefit from multi-spectral imaging.

\citet{Bargoti2017b} also employed the Faster
R-CNN architecture for fruit detection. They produced datasets from
images captured in orchards by a robotic ground vehicle for apples and
mangoes, and a dataset for almonds, also in orchards, but using a
hand-held DSLR camera (digital single-lens reflex). Employing image augmentation
strategies on training, the authors reached $F_1$-scores up to 0.90 for mangoes and apples 
and 0.77 for almonds. A surprising result reported by the authors is that
transfer learning between farms (same crop) or between orchards of
different crops showed little advantage compared to ImageNet transfer
learning. Bargoti and Underwood state such result increase the body
of evidence showing ImageNet features applicability for a broad range
of tasks.

In Faster R-CNN, detection is performed in two stages. The first stage
uses a \emph{region proposal network}, an attention mechanism
developed as an alternative to the earlier sliding window based
approaches. In the second stage, bounding box regression and object
classification are performed. Faster R-CNN is fairly recognized as a
successful architecture for object detection, but it is not the only
\emph{meta-architecture} \citep{Huang2017} able to reach
state-of-the-art results. Another group of architectures is the
\emph{single shot detector} (SSD) meta-architecture \citep{Huang2017,
  Liu2016}, single feed-forward convolutional networks able to 
predict classes and bounding boxes in a single stage. The YOLO
(\emph{You Only Look Once}) networks, proposed by Redmon and Farhadi
\citep{Redmon2016, Redmon2017}, are examples of the SSD family. 

Grape clusters present larger variability on size, shape and 
compactness compared to other fruits like peppers, apples or mangoes 
\citep{Bargoti2017b, Sa2016}. A focus on berry detection, such as in
\citet{Nuske2011, Nuske2014}, can be seen as a way to
circumvent grape cluster variability, performing yield prediction
over berry counting, consequently bypassing the grape cluster segmentation
problem. CNNs can learn representations of complex 
visual patterns \citep{Goodfellow-et-al-2016}, so are an interesting
alternative for grape cluster detection. However, object detection using
bounding boxes could be insufficient for yield prediction
applications, considering the enormous variability in grape clusters'
shapes and compactness. On the other hand, semantic segmentation (the
classification of pixels as fruit or background) could also be
inadequate, considering the severe occlusion between fruits observed in
orchards \citep{Bargoti2017b}. \emph{Instance segmentation}
(Figure~\ref{fig:dataset-example}), the combined task of object
detection (where are the grape clusters?) and pixel classification
(this pixel belongs to which cluster?), is an alternative machine learning task
formulation for yield prediction and automated harvesting applications.

Mask R-CNN~\citep{He_2017_ICCV} is a derivation of Faster R-CNN
able to perform instance segmentation, jointly
optimizing region proposal, bounding box regression and semantic pixel
segmentation. However, differently of object detection in which
rectangular bounding boxes annotations are sufficient for training,
instance segmentation needs image pixels to be properly attributed to
an instance or to the background in the training dataset for
supervised learning. In Section~\ref{sec:Methodology}, we describe a methodology for fruit
instance segmentation based on Mask R-CNN, including a
novel instance annotation tool for objects of complex shape. We
compare YOLO and Mask R-CNN results on wine grape cluster detection,
and we evaluate Mask R-CNN results on cluster instance segmentation.  

Fruit detection in single images can be the \emph{perceptual step} in a fruit
counting system, but without some sort of integration of the information
produced for the orchard, accurate prediction of yield is not
possible. \citet{Liu2019}, extending the work in
\citet{Bargoti2017b}, integrated the fruit
detection results in image sequences (video frames) performing
\emph{object tracking}. Employing the bounding box centers as
observations, the authors implemented an object tracker based on the
Kanade-Lucas-Tomasi algorithm (optical flow), Kalman filters and the
Hungarian Assignment algorithm, tracking fruits in video frame
sequences. To address issues caused by missing detections and
occlusions, they performed \emph{structure-from-motion},
recovering three-dimensional information using the box centers and their
inter-frame correspondence. Associating fruit locations in 3-D and the
CNN detection in 2-D frames, \citet{Liu2019} integrated data from a
camera moving along a mango orchard row, avoiding double-counting from
the same fruit observed in different frames, addressing occlusions and
localizing yield information in space.  Similarly, we propose a simple but effective \emph{spatial
  registration step} for fruit tracking and counting, also employing
3-D association from structure-from-motion data.      

\section{Materials and methods}
\label{sec:Methodology}

The proposed methodology introduces a new public dataset for image-based 
grape detection, including a novel method for interactive mask annotation for 
instance segmentation (Section~\ref{sec:TheDataset}). Three neural networks are
trained and evaluated for fruit detection: Mask R-CNN \citep{He_2017_ICCV}, 
YOLOv2 \citep{Redmon2016} and YOLOv3 \citep{Redmon2018} (Section~\ref{sec:CNNs}). 
Evaluation measures for semantic segmentation, object detection, and instance 
segmentation variants are presented in Section~\ref{sec:Eval}. Section~\ref{sec:SpatialReg} 
presents our approach for spatial integration.

\subsection{The dataset}
\label{sec:TheDataset}

The \emph{Embrapa Wine Grape Instance Segmentation Dataset} (WGISD) is
composed by 300 RGB images showing 4,432
grape clusters from five different grape varieties, as summarized in
Table~\ref{table:GenInfoData}. {All images were captured from a single
winery, that employs \emph{dual pruning}: one for shaping (after
previous year harvest) and one for production, resulting in canopies
of lower density. No pruning, defoliation or any intervention in the
plants was performed specifically for the dataset construction: the
images capture a real, trellis system-based wine grape production. The
camera captures the images in a frontal pose, that means the camera
principal axis is approximately perpendicular to the wires of the
trellis system and the plants rows. As seen in
Table~\ref{table:GenInfoData}, the \emph{Syrah} images were taken in a
different field visit, one year earlier, and consist a smaller set if
compared to the other four varieties.} \ref{appendix:WGISD} presents a
detailed description for the dataset, following the guidelines
proposed by \citet{Gebru2018} for dataset characterization, and
including information about cameras, field location, pre-processing
and file formats. The WGISD is publicly available \citep{ZenodoWGISD}
under the CC BY-NC 4.0 (Attribution-NonCommercial 4.0 International)
license. 

\begin{table}
  \footnotesize
  \caption{General information about the dataset: the grape
    varieties and the associated identifying prefix, the date of
    image capture on field, number of images (instances) and the
    identified grapes clusters.}
  \begin{tabular}{llrrrr}
    \hline
    Prefix & Variety & Date & Images & Boxed clusters & Masked clusters\\
    \hline
    CDY & \emph{Chardonnay} & 2018-04-27 & 65 & 840 & 308\\
    CFR & \emph{Cabernet Franc} & 2018-04-27 & 65 & 1,069 & 513\\
    CSV & \emph{Cabernet Sauvignon} & 2018-04-27 & 57 & 643 & 306\\
    SVB & \emph{Sauvignon Blanc} & 2018-04-27 & 65 & 1,317 & 608\\
    SYH & \emph{Syrah} & 2017-04-27 & 48 & 563 & 285\\
    \hline
    Total &  &  & 300 & 4,432 & 2,020\\
    \hline
  \end{tabular}
  \label{table:GenInfoData}
\end{table}

To be employed on supervised instance segmentation training, WGISD
has to provide a set of \emph{masks} that properly segment grape
clusters. {WGISD provides binary masks for 2,020 clusters from the
4,432 total, as seen in Table~\ref{table:GenInfoData}. Mask
annotation for instance segmentation 
is a laborious task that requires custom tools to allow the annotation
of hundreds of images in 
limited time. The VGG Image Annotator (VIA) \citep{Dutta2016} is a
popular tool used by the computer vision community.} It allows users to
mark objects of interest using rectangles, circles, ellipses or
polygons. In an interesting attempt to automatize annotation,
\citet{Acuna2018} proposed an interactive 
tool that uses a neural network (Polygon-RNN++) to predict the next
vertex in polygonal annotations.

In WGISD construction, the accurate annotation of complex objects
in natural scenes using polygonal shapes proved to be extremely
laborious, even when employing the vertex prediction facilities from
Polygon-RNN++. To relieve the annotation process, we have created an
annotation tool based on interactive image segmentation by graph
matching, as proposed by \citet{Noma2012}. This
method starts from an over-segmentation, produced by the watershed
algorithm \citep{vincent1991watersheds}, to create an
\emph{attributed relational graph} (ARG) representation for the image --
$G_i$. Then, the user can freely mark the image using
\emph{scribbles}. Such marks are used to create a \emph{model graph}
$G_m$, a labeled ARG. Exploiting the spatial relations among ARGs
vertices, a match is computed between the model graph $G_m$ and the
input image graph $G_i$, allowing the propagation of labels from $G_m$
to $G_i$.

\begin{figure}
  \centering
  \begin{tabular}{ccc}
    \includegraphics[width=0.3\textwidth]{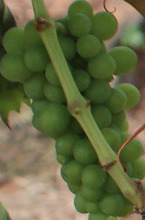} &
    \includegraphics[width=0.3\textwidth]{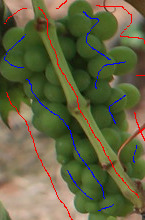} &
    \includegraphics[width=0.3\textwidth]{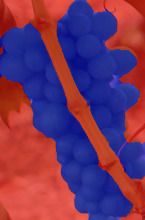} \\
    (a) & (b) & (c)
  \end{tabular}
  \caption{Instance annotation using interactive image segmentation by
  attributed relational graphs. (a) Grape cluster delimited using
  a standard bounding box annotation. (b) Scribbles drawn by the user
  (blue for grapes, red for background or foreground structures). (c)
  Segmentation produced by the graph matching procedure.}
  \label{fig:Annot}
\end{figure}

Figure~\ref{fig:Annot} shows an example of grape annotation. The
dataset was previously annotated for object detection using standard
rectangular bounding boxes (see \ref{appendix:WGISD} for
details). The instance annotation tool uses the bounding boxes as
inputs, displaying each grape cluster for an interactive image
segmentation procedure by graph matching (Figure~\ref{fig:Annot}~(a)). An 
annotator can draw scribbles, freely
marking pixels that should be considered part of the grape cluster and
pixels that are part of the background or occluding foreground
objects (Figure~\ref{fig:Annot}~(b)). The graph matching-based
algorithm uses the scribbles to produce a segmentation,
propagating the labels from the model to the input image
(Figure~\ref{fig:Annot}~(c)). The tool allows the scribble marking and graph matching
steps to be repeated by the user until a reasonable annotation is
achieved. Finally, the grape pixels are stored as masks for supervised
instance segmentation learning. Readers interested in a detailed
description of the graph matching algorithm should refer to
\citet{Noma2012}.

\subsection{The perceptual step: CNN architectures}
\label{sec:CNNs}

\begin{figure}
  \centering
  \begin{tabular}{cc}
    \includegraphics[width=0.45\textwidth]{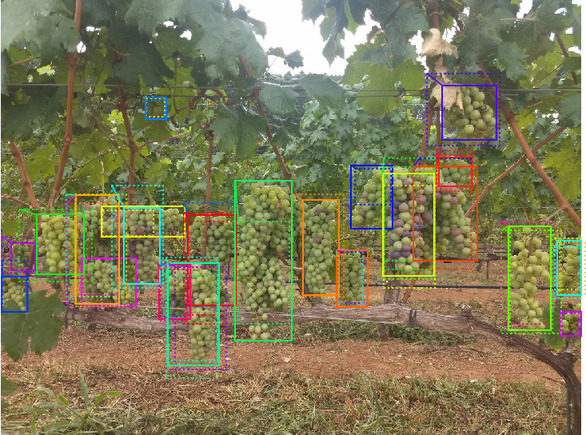}
    & \includegraphics[width=0.45\textwidth]{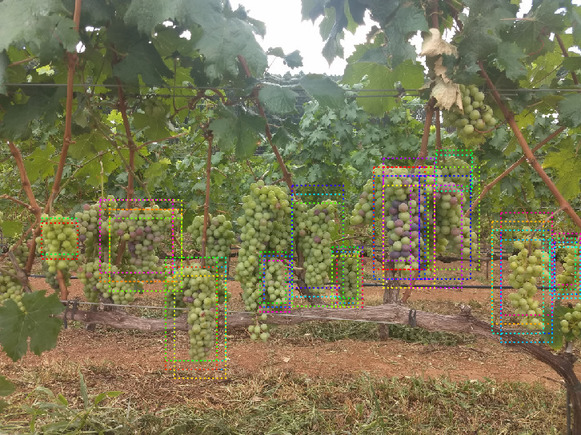}\\
    (a) & (b)\\
    \includegraphics[width=0.45\textwidth]{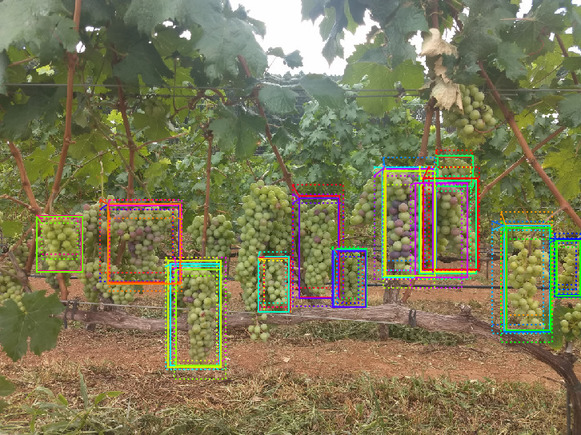}
    & \includegraphics[width=0.45\textwidth]{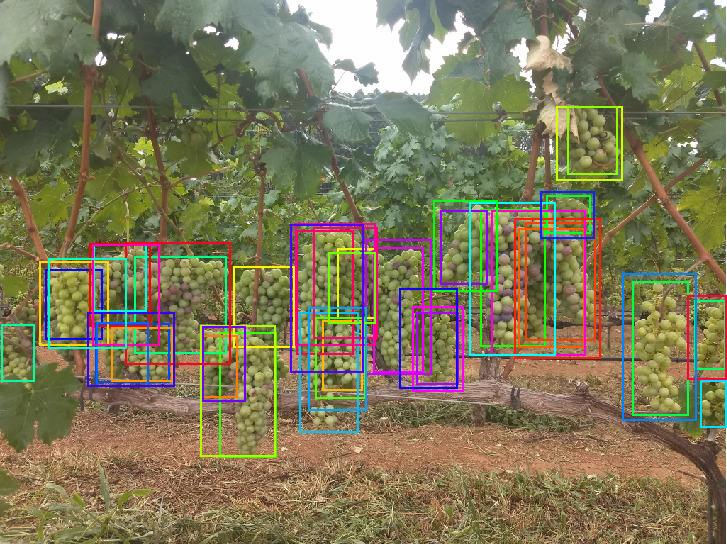}\\
    (c) & (d)
  \end{tabular}
  \caption{RPN network under action. (a) Targets for training the RPN,
    built from the training set - note the anchors (dashed lines) and
    the location and size deltas (solid lines). (b) Subset of the top
    rated anchors (few anchors shown to improve visualization for the
    reader). (c) Subset of the top anchors, after refinement. (d)
    Final regions found by the RPN for the image after non-max
    suppression.} 
  \label{fig:RPN}
\end{figure}

\begin{figure}
  \centering
  \begin{tabular}{ll}
    \includegraphics[width=0.45\textwidth]{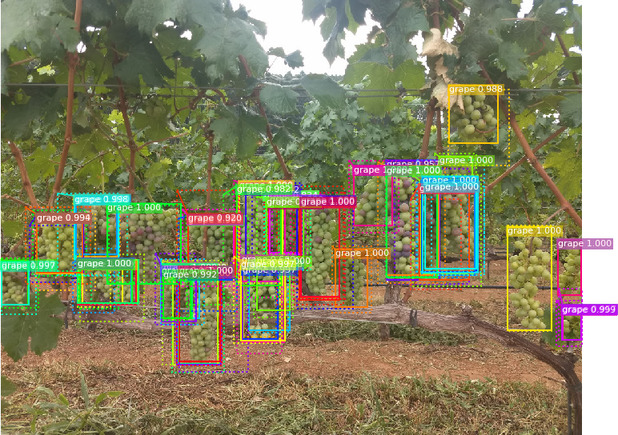}
    & \includegraphics[width=0.45\textwidth]{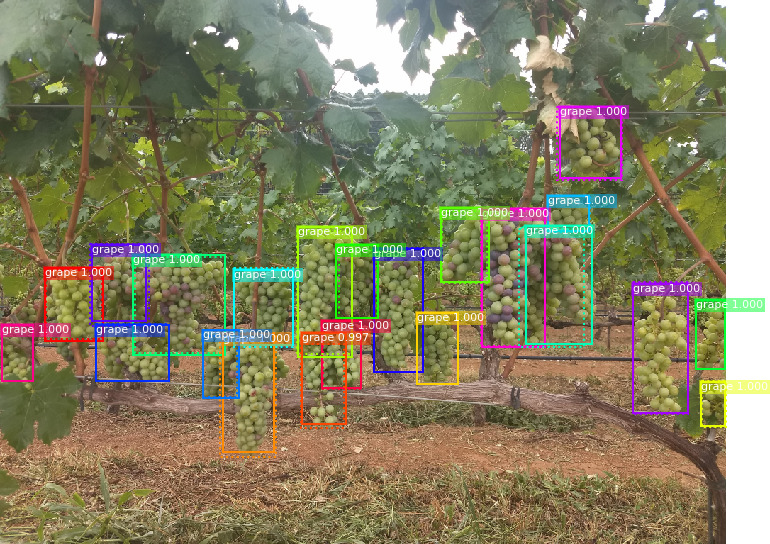}\\
    \multicolumn{1}{c}{(a)} & \multicolumn{1}{c}{(b)}\\
    \includegraphics[width=0.425\textwidth]{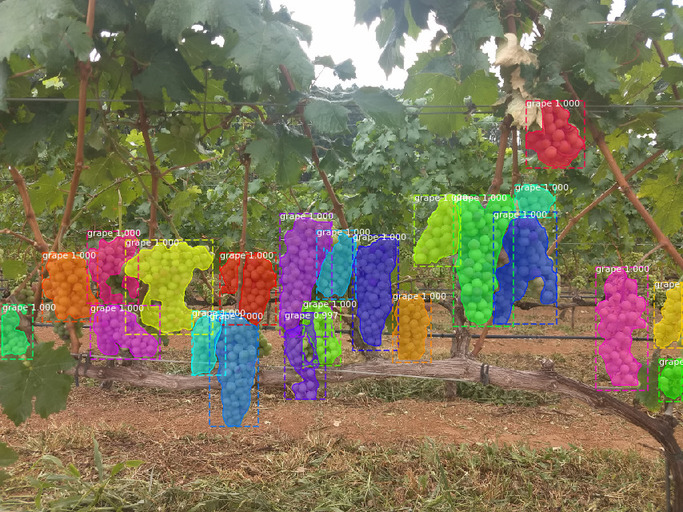}
    & \includegraphics[width=0.425\textwidth]{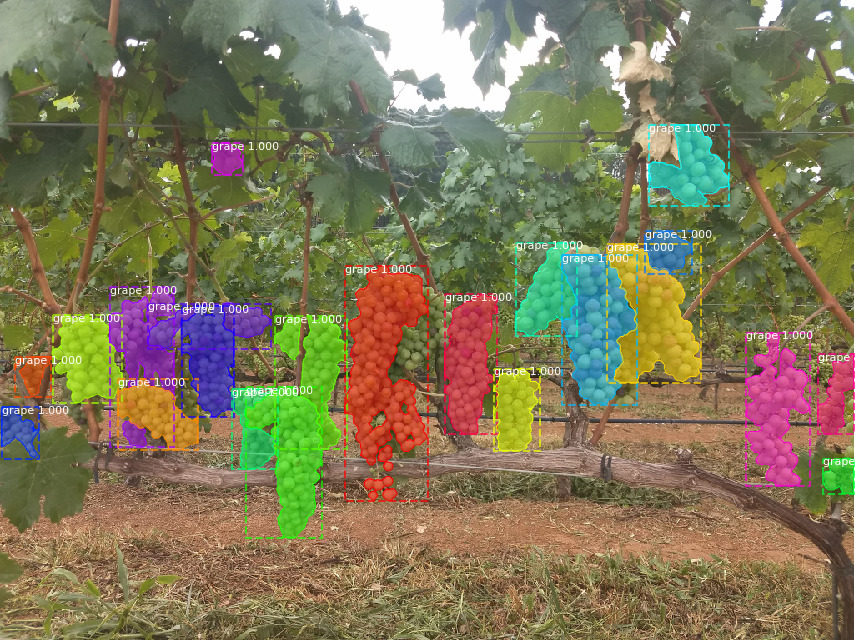}\\
    \multicolumn{1}{c}{(c)} & \multicolumn{1}{c}{(d)}\\
  \end{tabular}
  \caption{Mask R-CNN under action. (a) Class-specific bounding box
    refinement. (b) Results after low-confidence filtering and non-max
    suppression. (c) Final results after FCN pixel classification (d)
    Ground-truth.}
  \label{fig:MRCNN}
\end{figure}

Mask R-CNN \citep{He_2017_ICCV} is a consolidation of a long sequence
of works developed by He, Dollár, Girshick and colleagues. This
network is essentially the combination of a Faster R-CNN object detector
\citep{NIPS2015_5638} and a \emph{fully convolutional network} (FCN)
\citep{Shelhamer2017} for semantic segmentation, providing a complete,
end-to-end, instance segmentation solution. The Faster R-CNN is also a
combination of two architectures: a \emph{region proposal network}
(RPN) and an object detector, the Fast R-CNN \citep{Girshick2015}. RPN
works as an attention mechanism, finding \emph{anchors} in the 
feature space, rectangular boxes that can contain objects of
interest (Figure~\ref{fig:RPN}). The Fast R-CNN is composed of a
softmax object classifier and a per-class bounding box regressor
(Figure~\ref{fig:MRCNN}). The Mask R-CNN employs as feature extractor
a \emph{feature pyramid network} (FPN) \citep{Lin_2017_CVPR}, an
architecture able to create semantic feature maps for objects at
multiple scales, built over a ResNet \citep{He2016}. 

Another approach to object detection is to predict the locations and 
the objects' class in a single step, in order to avoid a previous region proposal procedure. 
\citet{Huang2017} refer to this approach as \emph{single shot detector meta-architecture}, and the YOLO networks proposed by 
\citet{Redmon2016, Redmon2017} are prominent members of this family.
In the YOLO networks, the image is split into a fixed grid of $S \times S$ \emph{cells}. 
A cell is responsible for performing a detection if an object center is over it. Each cell
is associated to $B$ boxes, composed by 5 values representing the object center $(c_x, c_y)$,
the object width and height and a \emph{confidence score} that represents the model
confidence that the box contains an object and also the accuracy of the box boundaries regarding
the object. The box also includes $C$ conditional class probabilities, one to each
class of objects. Consider, for example, a $7 \times 7$ grid of cells ($S = 7$), where
each cell predicts $B = 2$ boxes for 20 different classes of object ($C = 20$). The YOLO
network will produce a $7 \times 7 \times 30$ output tensor. This means a $B \cdot 5 + C$ vector
for each one of the 49 cells. The training step tries to minimize a loss function defined over
such a tensor, performing detection and classification in a single
step. The YOLOv2 and YOLOv3 networks have a few differences, mainly
regarding their feature extraction convolution part. YOLOv3 
presents a deeper convolutional network that incorporate some
state-of-the-art techniques such as residual networks \citep{He2016},
skip connections and multi-scaling (similar to FPNs). YOLOv3
classification is based in multi-label classification instead of the
softmax employed by YOLOv2, allowing the former able to deal with
multi-class problems. 

\subsection{Training}
\label{sec:Training}

{Table~\ref{table:VarsSplits} shows the splitting between
training/validation and test sets. The images were assigned randomly
to each set in an 80-20\% proportion. It is important to note that
{\it Cabertnet Franc} and {\it Sauvignon Blanc} varieties presented a
higher number of clusters per image. In Section~\ref{sec:Results}, we
will show that although the differences in the numbers of images and
clusters, the results are very similar for all five grape varieties.}
For instance segmentation, a set of 110 images presenting masks is
available for training. We have split it into an 88 images training
set (1,307 clusters) and a validation set composed of 22 images (305 
clusters).

\begin{table}
{
  \caption{{Training and test sets sizes, stratified by grape
    variety. Images were randomly selected for each set in an
    80-20\% split (approximately). In the \emph{Masked clusters}
    column, the number in parentheses corresponds to the number of
    \emph{masked} images available.}}  
  \footnotesize
  \begin{tabular}{llrrr}
    \hline
    & Variety & Images & Boxed clusters & Masked clusters\\\hline
    &	CDY 	& 50 	& 660 & 227 (18)\\
    & CFR 	& 55 	& 910 	& 418 (27)\\
    Train./Val.      & CSV 	& 48 	& 532 	& 241 (23)\\
    & SVB 	& 51 	& 1034 	& 502 (24)\\
    & SYH 	& 38 	& 446 	& 224 (18)\\\cline{2-5}
    & Total     & 242   & 3582  & 1612 (110)\\\hline    
    &	CDY 	& 15 	& 180 	& 81 (6)\\
    & CFR 	& 10 	& 159 	&  95 (6)\\
    Test  & CSV 	& 9     & 111   &  65 (5)\\
    & SVB 	& 14   	& 283   & 106 (5)\\
    & SYH 	& 10 	& 117 	&  61 (5)\\\cline{2-5}
    & Total   & 58    & 850 & 408 (27)\\
    \hline
  \end{tabular}
  \label{table:VarsSplits}
}
\end{table}

\begin{figure}
  \centering
  \begin{tabular}{llll}
    {Original Image}\\
    \includegraphics[width=0.24\textwidth]{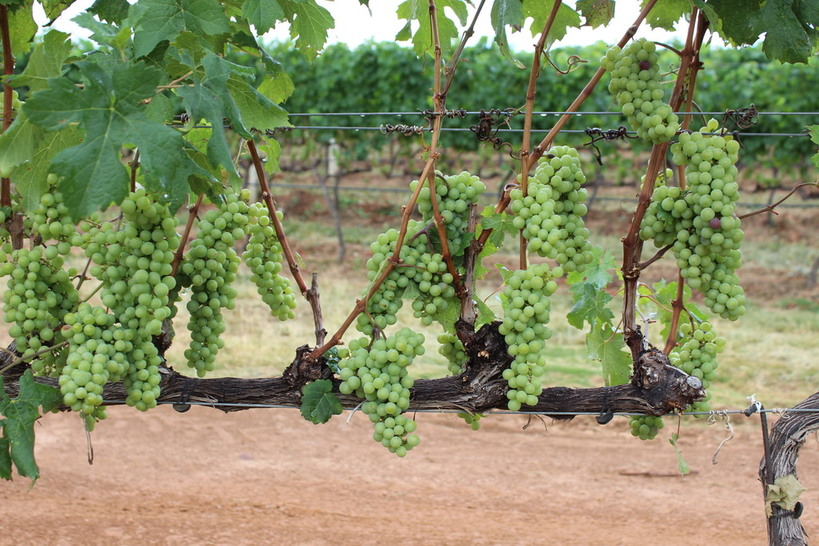}\\
    {Augmentations}\\
    \includegraphics[width=0.24\textwidth]{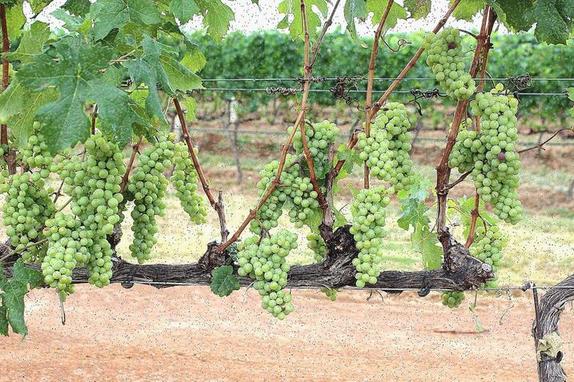} &
    \includegraphics[width=0.24\textwidth]{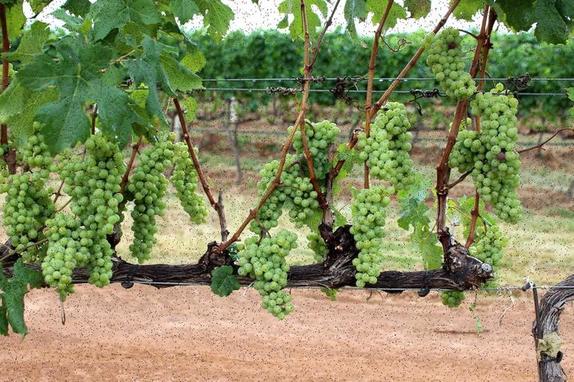} &
    \includegraphics[width=0.24\textwidth]{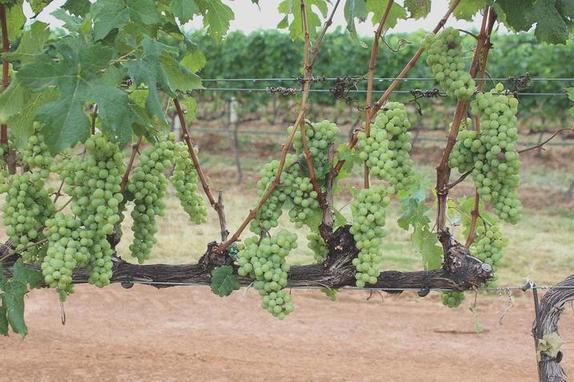} &
    \includegraphics[width=0.24\textwidth]{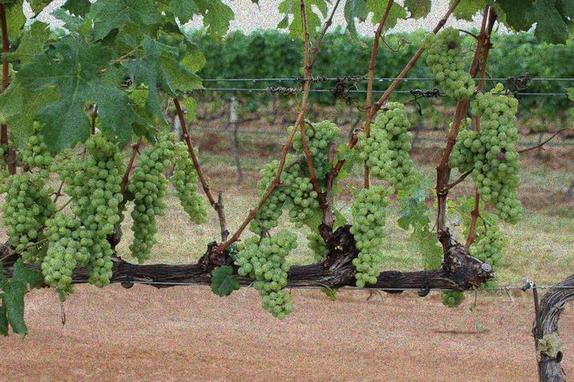}\\
    \includegraphics[width=0.24\textwidth]{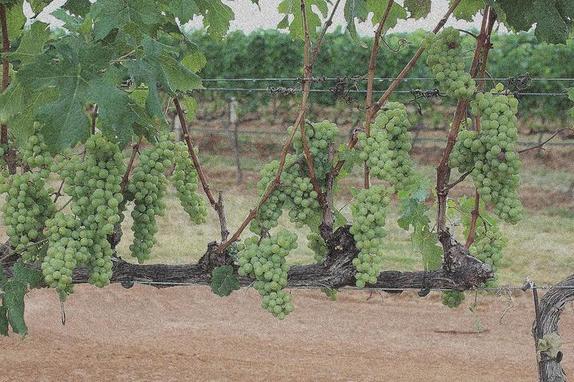} &
    \includegraphics[width=0.24\textwidth]{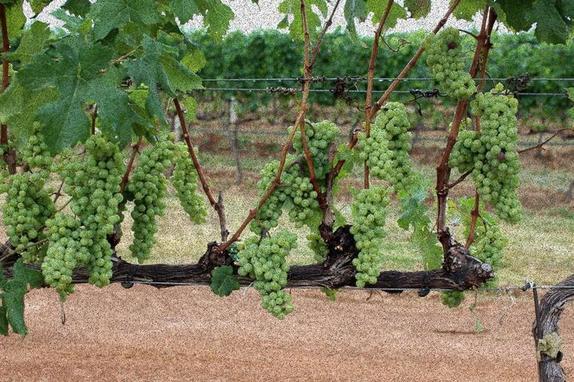} &
    \includegraphics[width=0.24\textwidth]{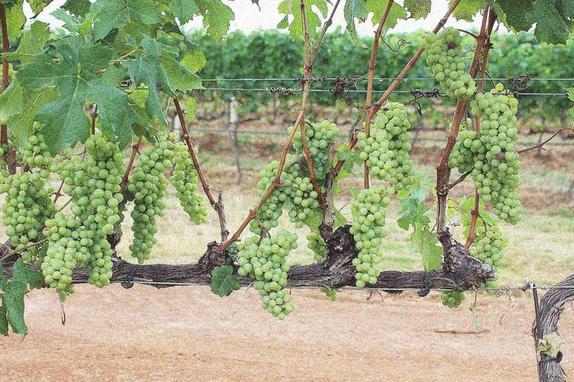} &
    \includegraphics[width=0.24\textwidth]{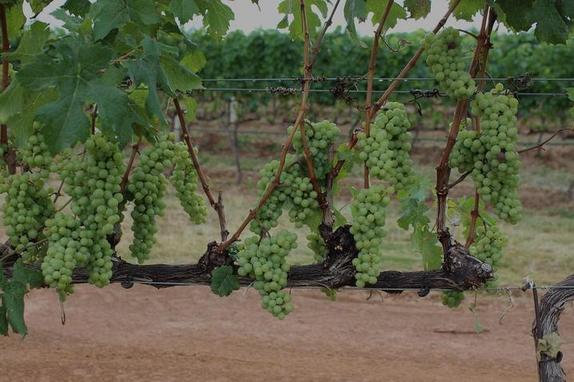}\\
    \includegraphics[width=0.24\textwidth]{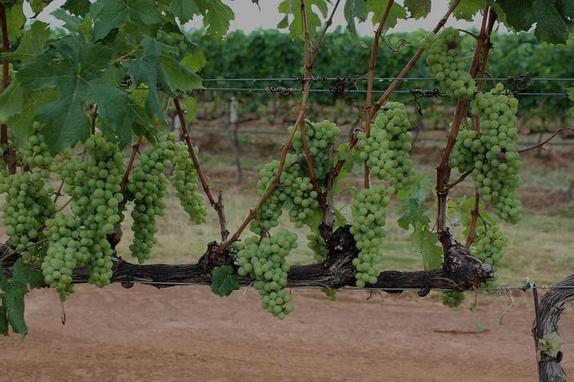} &
    \includegraphics[width=0.24\textwidth]{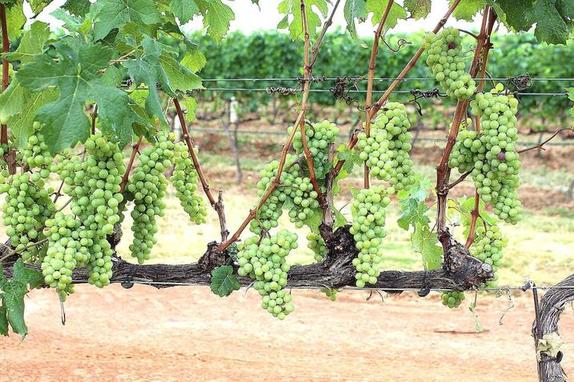} &
    \includegraphics[width=0.24\textwidth]{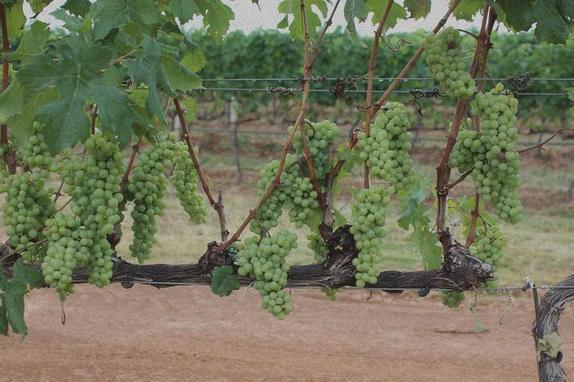} &
    \includegraphics[width=0.24\textwidth]{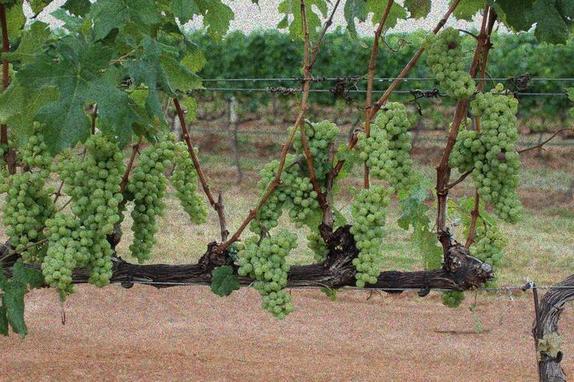}\\
    \includegraphics[width=0.24\textwidth]{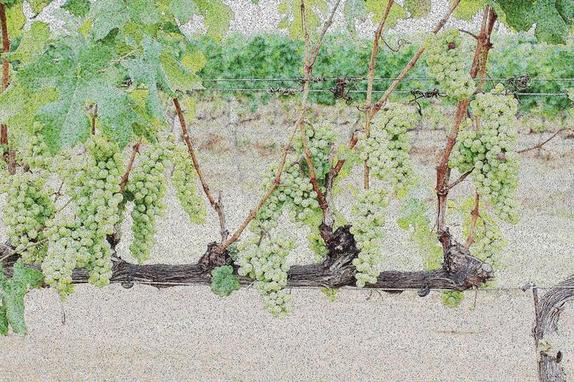} &
    \includegraphics[width=0.24\textwidth]{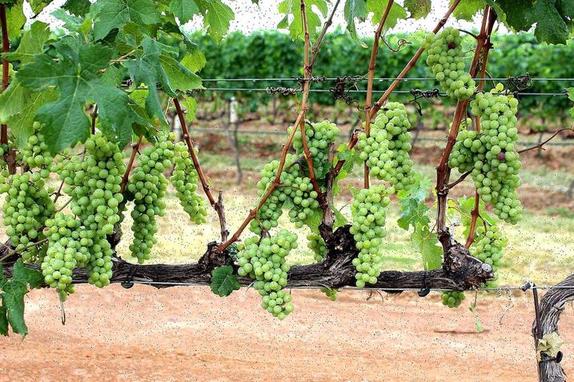} &
    \includegraphics[width=0.24\textwidth]{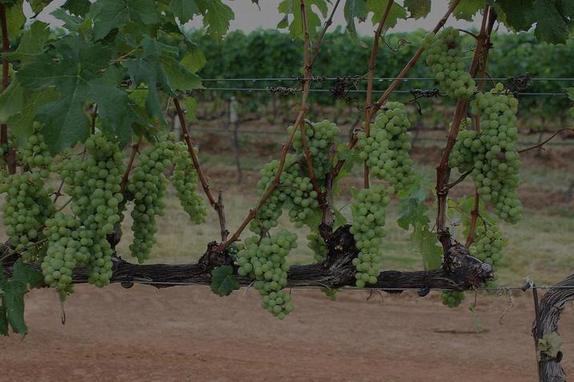} &
    \includegraphics[width=0.24\textwidth]{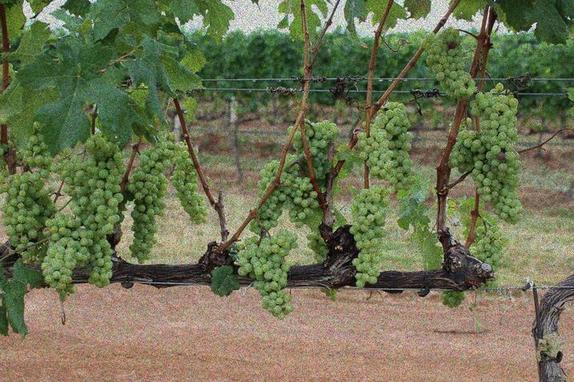}\\
    \includegraphics[width=0.24\textwidth]{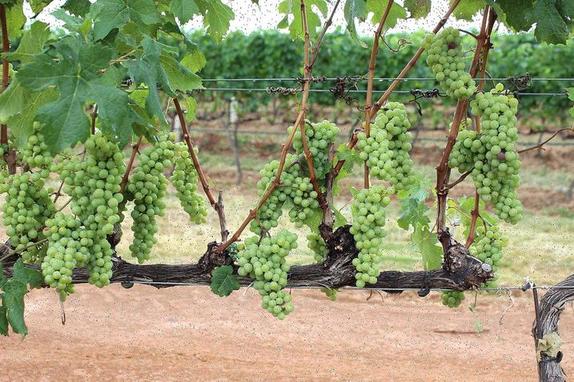} &
    \includegraphics[width=0.24\textwidth]{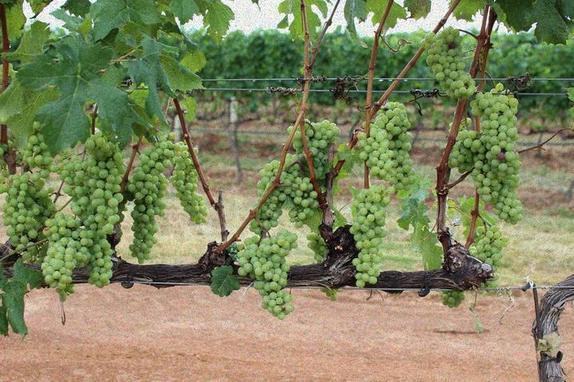} &
    \includegraphics[width=0.24\textwidth]{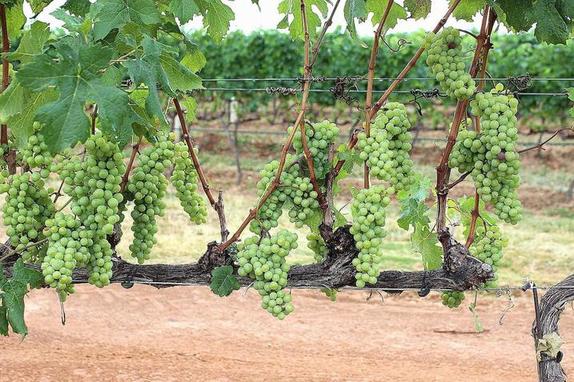} &
    \includegraphics[width=0.24\textwidth]{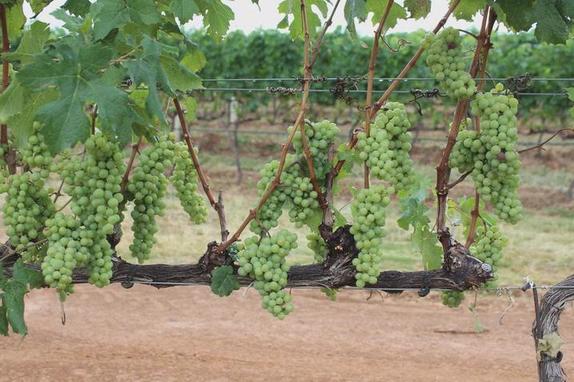}\\
  \end{tabular}
  \caption{{Image augmentations produced for a {\it Cabernet Franc}
    image. The randomized pipeline of transformations produces
    variations in lighting, contrast and noise. Dirty lenses are
    emulated using pixel dropouts.}} 
  \label{fig:Aug}
\end{figure}

We employed image augmentation to mitigate overfitting
\citep{Chollet2017}, adopting transformations of the original images
that could simulate field conditions: differences in lighting, camera
focus, noise or dirty lenses. We applied Gaussian blur, contrast
normalization, additive Gaussian noise and pixel
dropouts\footnote{Similar to ``pepper'' noise -- see \texttt{imgaug} 
  documentation for details \citep{imgaug}.} using the \texttt{imgaug} library
\citep{imgaug}. These augmentations were randomly selected and ordered
in such a way that \emph{different transformations} were applied for each
source image. {We have applied 20 random augmentations for each image,
as shown in Figure~\ref{fig:Aug}, which produced a rich set of
variations, potentially reflecting real field conditions.}

We employed the Keras/TensorFlow-based implementation for Mask R-CNN
developed by Matterport, Inc., publicly available at GitHub
\citep{Matterport}. The network was initialized using the weights
previously computed for the COCO Dataset \citep{MSCOCO}. No layer was
frozen during training, so all weights could be updated by the
training on the grapes dataset. Due to GPU memory limitations, to
allow multiple images per batch, input images are resized to $1024
\times 1024 \times 3$ tensors, preserving aspect ratio by applying
zero padding as needed.
Two feature extraction architectures were evaluated:
ResNet-101 and the shallower ResNet-50 \citep{He2016}. For the YOLO networks, we employed 
Darknet, the original implementation developed by \citet{Redmon2016}, initialized using
pre-trained weights from ImageNet \citep{Deng2009ImageNet}. In our single-class grape 
detection case, $C = 1$. {Training was performed on a computer
containing a single NVIDIA TITAN Xp GPU (12~GB memory), an Intel
i7-x990 CPU and 48 GB RAM, and running Ubuntu~18.04~LTS. For
Mask~R-CNN, training was performed in approximately 10~hours (100
epochs, around 6 minutes per epoch). YOLO training (v2 and v3) spent four days 
using the same hardware.}

\subsection{Evaluation}
\label{sec:Eval}

The WGISD dataset allows evaluations for the semantic segmentation,
object detection and instance segmentation problems. This work will
present results using the standard metrics of \emph{precision} ($P$),
\emph{recall} ($R$), and their harmonic mean ($F_1$), as usual in the
information retrieval literature:
\begin{equation}
  P = \frac{N_{\mathrm{tp}}}{N_{\mathrm{tp}} + N_{\mathrm{fp}}},
\end{equation}
\begin{equation}
  R = \frac{N_{\mathrm{tp}}}{N_{\mathrm{tp}} + N_{\mathrm{fn}}}, \; \textrm{and} 
\end{equation}
\begin{equation}
  F_1 = \frac{2 \cdot P \cdot R}{P + R}.
\end{equation}
These measurements depend on the
number of \emph{true positives} ($N_{\mathrm{tp}}$), \emph{false negatives} ($N_{\mathrm{fn}}$)
and \emph{false positives} ($N_{\mathrm{fp}}$), which need to be properly defined for
each type of problem:
\begin{itemize}
  \item For \emph{semantic segmentation}, we are considering just one class
  (grape) and pixel classification. In this case, we employ the
  masked images in the test set for evaluation, where the grape pixels are
  properly marked (27 images, 408 grape clusters). $N_{\mathrm{tp}}^\mathrm{seman}$ is the
  number of  pixels correctly classified  as grape pixels according to
  the ground truth, $N_{\mathrm{fn}}^\mathrm{seman}$ the number of grape pixels incorrectly
  classified as non-grape pixels, and $N_{\mathrm{fp}}^\mathrm{seman}$ the number of
  non-grape pixels wrongly reported as grape ones by the classifier. Such three measures allow
  the computation of $P_{\mathrm{seman}}$ and $R_{\mathrm{seman}}$, respectively precision and
  recall, for the semantic segmentation problem.
\item In \emph{object detection}, each grape cluster instance is
  localized by a rectangular bounding box. A hit or a miss is
  defined by a one-to-one correspondence to ground truth instances,
  obeying an \emph{intersection over union} (IoU) threshold computed
  using the rectangular areas and their intersections. $N_{\mathrm{tp}}^\mathrm{box}$ is the
  number of correctly predicted instances (bounding boxes) and $N_{\mathrm{fn}}^\mathrm{box}$
  and $N_{\mathrm{fp}}^\mathrm{box}$ are similarly defined. These three
  measures give the values for $P_\mathrm{box}$ and $R_\mathrm{box}$, respectively precision and
  recall, for the object detection problem and the evaluation is
  performed for the entire test set (58 images, 837 grape clusters).       
 \item \emph{Instance segmentation} follows the same instance-based
   logic as in object detection, but IoU is computed using the areas
   and intersections of the \emph{masks} instead of rectangular bounding
   boxes. Again, we are limited to the masked images in the test set: 27
   images containing 408 clusters. The measures are $P_{\mathrm{inst}}$ and $R_{\mathrm{inst}}$ for
   instance segmentation precision and recall, respectively. 
\end{itemize}
Mask R-CNN results can be evaluated for the three problems, but the
YOLO-based results are just evaluated regarding object detection. 

\subsection{Spatial registration: 3-D association}
\label{sec:SpatialReg}

Structure-from-Motion (SfM) \citep{HZ2003} is a fundamental achievement in
computer vision and a core component in modern photogrammetry. It solves
the camera pose and scene geometry estimation simultaneously,
employing only image matching and bundle adjustment \citep{Triggs2000}, and
finding three-dimensional structure by the motion of a single
camera around the scene (or from a set of independent cameras). In a
previous work \citep{Santos2017}, we showed that SfM can
recover vine structure from image sequences on vineyards. It is an
interesting alternative to integrate image data from different camera poses
registering the same structures in space. Similarly to Liu~{\it et
  al.}~\citep{Liu2019}, we use 3-D data from the COLMAP SfM
software \citep{Schoenberger2016} to perform spatial registration, integrating the fruit
instance segmentation data produced by the perceptual CNN-based step. 

Consider the directed graph $G = (V, E)$, where $V$ is a set of nodes $u_{i,j}$
 representing the $j$-th instance found by the neural network in the $i$-th
frame. Consider the set of $\mathcal{X} = \{\mathbf{X}_k\}_{k=1..M}$ of
$M$ three-dimensional points $\mathbf{X}_k$ found by the SfM 
procedure. We create an oriented edge $(u_{i,j}, v_{i',j'}) \in E$,
considering $i < i'$,  if there is a 3-D point $\mathbf{X}_k$ that
projects to the instance $j$ in frame $i$ and to instance $j'$ in frame
$i'$. In other words, there is a link between instances from two
different frames if there is a three-dimensional point whose
2-D projections are contained in the masks associated to 
these instances, evidence they could be observing the same object
in the 3-D world.

\begin{figure}
  \centering
  \includegraphics[width=\textwidth]{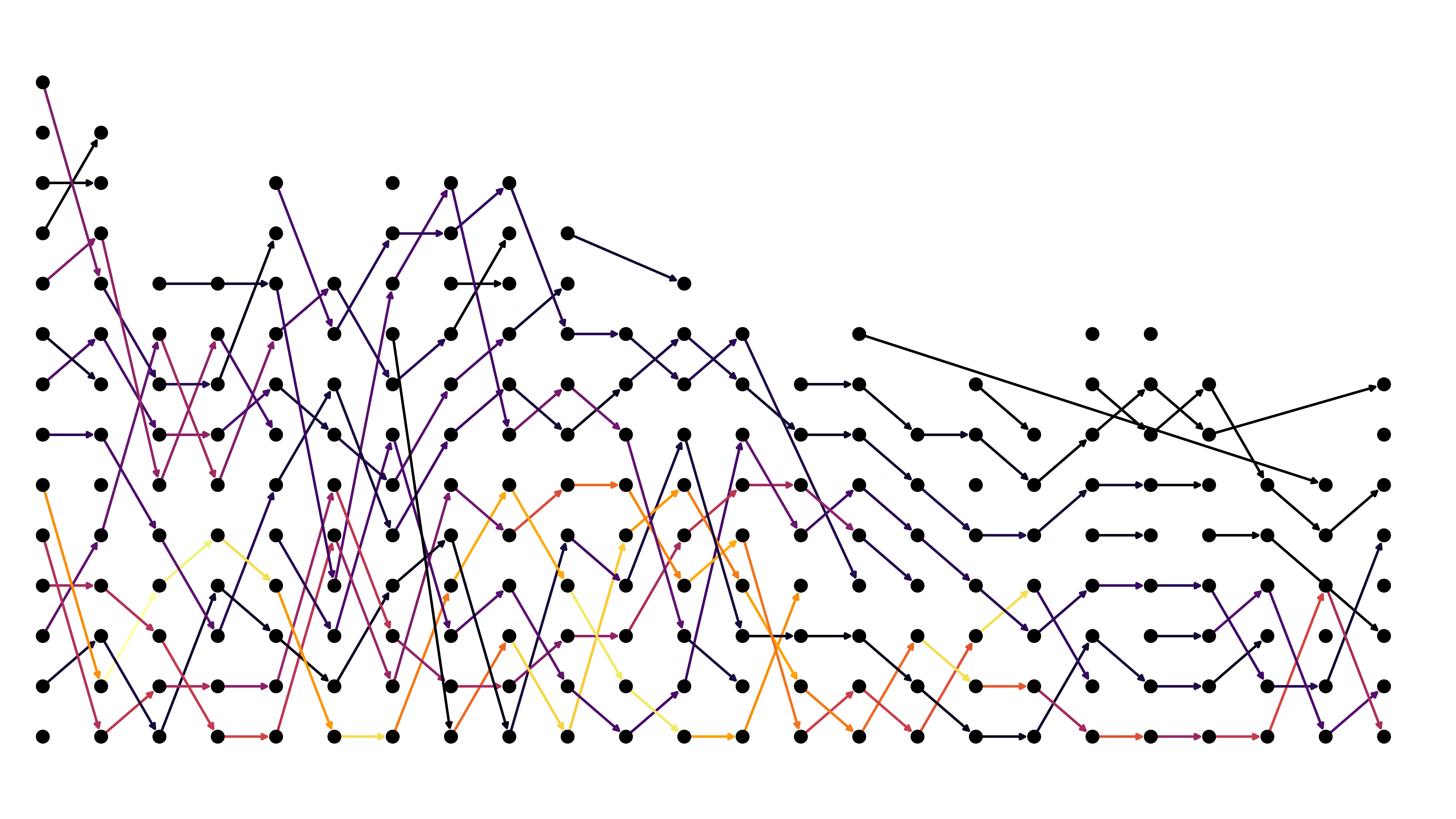}\\
(a)\\
\includegraphics[width=\textwidth]{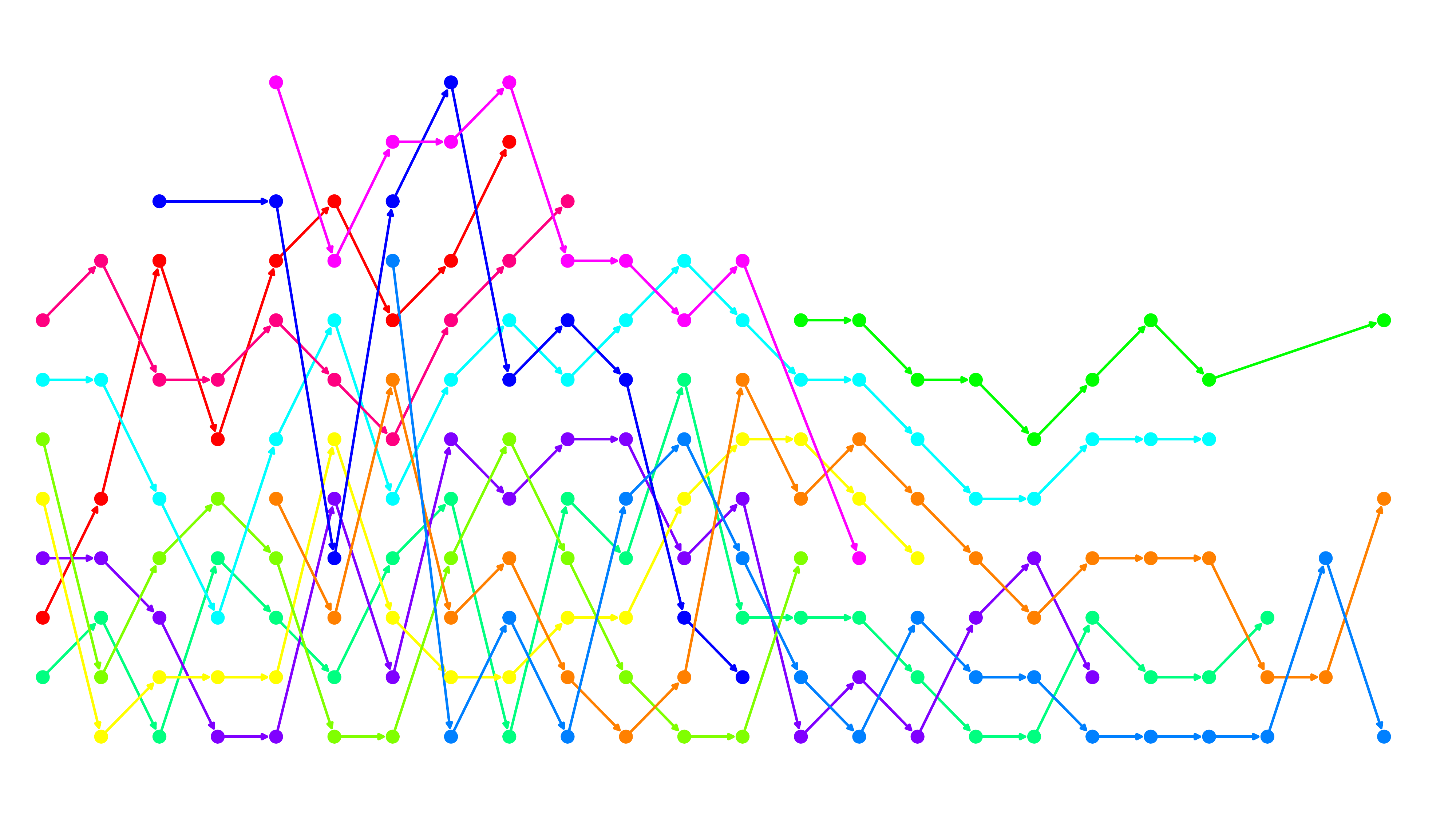}\\
(b)
  \caption{Instance matching and tracking using 3-D assignment. Each
    column represents the instances found by a neural network in a
    video frame. (a) Matching found by 3-D assignment -- the edges are
    strong (warm colors) as the number of 3-D points linking
    the instances. (b) Instance tracking found by looking for the
    deepest paths in the graph -- each color represents an individual
    grape cluster.} 
  \label{fig:InstanceMatching}
\end{figure}

{Figure~\ref{fig:InstanceMatching}~(a) shows the graph $G$. Each
node represents an instance (a grape cluster) and a column of nodes
represents all instances found in a frame $i$. Neighboring columns
represent successive frames $i$ and $i'$, $i < i'$. $G$ is a
directed graph, with oriented edges indicating instance association
between frames.} Each edge has a \emph{weight} $w[u_{i,j}, v_{i',j'}]$ that
indicates the total number of three-dimensional points that links the two
instances, accumulating the evidence that associates instance $j$ in
frame $i$ to the instance $j'$ in $i'$.

The structure of $G$ is affected by occlusions: if changes in camera
pose make two or more clusters to occlude each other, then two or more edges
will be incident to a node $u_{i,j}$. Similarly, when the camera movement
reveals two or more occluded clusters, two or more edges will flow
from the same node. We filter the edges in $E$ in such 
a way that, for each node, there is up to one incident edge and up to
one departing edge. The filtering strategy is simple: the
\emph{heaviest} (maximum weight $w$) edge is kept. The intuition
behind this strategy is it would favor the \emph{occluding} grape
cluster while the \emph{occluded} one is tracked by an edge spanning many
frames -- that means $(u_{i,j}, v_{i',j'})$ where $i' > i +
1$. These edges spanning many frames also help with the relocalization
of grapes occluded by other objects in the scene (leaves, trunks,
etc.) missed by the neural network in some frames. 

{After edge filtering, nodes are sorted by their frame index $i$ and,
for each node $u_{i,j}$, we find the longest path in $G$ using
depth-first search on edges, corresponding to the track of one grape
cluster along the frame sequence. Too short paths (we use a threshold
of 5 edges) are filtered out, an attempt to remove false positives
from the perceptual stage by integrating evidence from multiple
frames. Figure~\ref{fig:InstanceMatching}~(b) illustrates the final
set of longest and disjoints paths, where different colors
discriminate different tracks (different grape clusters). The number
of paths is an estimation of the total number of grape clusters in the
entire image sequence.}

\section{Results}
\label{sec:Results}

The validation set was employed to select the best models for further evaluation
on the test set. For the Mask~R-CNN, the ResNet~101 feature extraction backbone
produced the best results. Table~\ref{table:InstanceSegmentation} presents the evaluation of 
predictions produced by Mask~R-CNN for instance segmentation, considering the masked test
{set (408 clusters in the ground truth) and confidence threshold of 0.9
for the grape class\footnote{{In the experiments, Mask~R-CNN did not exhibit great
  sensibility to the confidence threshold. We have tested 0.5, 0.7,
  0.9 and 0.95, all presenting very similar results, and $F_1$ variations
  inferior to  0.005.}}. The table shows the precision and recall measures}
for seven different values of IoU, from 30\% to 90\%. The corresponding values 
for $F_1$ score and \emph{average precision}\footnote{The AP
  summarizes the shape of the precision/recall curve, and it is
  defined as the mean precision at a set of equally spaced
  recall levels. See \citet{PascalVOC} for details.} (AP) as defined in Pascal VOC Challenge
\citep{PascalVOC} are also presented.  {Considering the diversity in
clusters sizes and shapes, IoU is specially important: higher values
indicate better grape berries coverage, a desirable property for yield
prediction applications. Lower IoU values indicate poorer berry
coverage or disagreement in clusters segmentation between prediction
and the ground truth, that means divergence in the berries' assignment
to clusters.}      

\begin{table}
  \caption{Instance segmentation results for Mask R-CNN. This
    evaluation was performed in the masked test set, considering a
    confidence level of 0.9 for the \emph{grape} class.}
  \footnotesize
  \begin{tabular}{rrrrr}
    \hline
    IoU & AP & $P_{\mathrm{inst}}$ & $R_{\mathrm{inst}}$ & $F_1$\\
    \hline
    0.3 & 0.855 & 0.938 & 0.892 & 0.915\\
    0.4 & 0.822 & 0.923 & 0.877 & 0.899\\
    0.5 & 0.743 & 0.869 & 0.826 & 0.847\\
    0.6 & 0.635 & 0.799 & 0.760 & 0.779\\
    0.7 & 0.478 & 0.696 & 0.662 & 0.678\\
    0.8 & 0.237 & 0.485 & 0.461 & 0.472\\
    0.9 & 0.008 & 0.070 & 0.066 & 0.068\\
    \hline
  \end{tabular}
  \label{table:InstanceSegmentation}
\end{table}

%
%
%
%
%
%
%

\begin{figure}
  \centering
  \begin{tabular}{clr}
    \space & Predicted & Ground Truth\\
    \rotatebox{90}{\small \it Chardonnay} &
    \multicolumn{2}{c}{\includegraphics[width=0.8\textwidth]{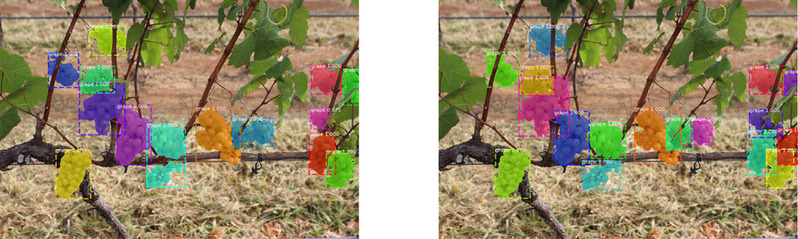}}\\
    \rotatebox{90}{\small \it Cabernet Franc} &
    \multicolumn{2}{c}{\includegraphics[width=0.8\textwidth]{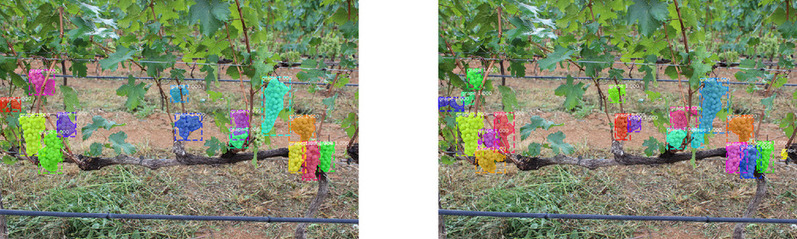}}\\
    \rotatebox{90}{\small \it Cabernet Sauvignon} &
    \multicolumn{2}{c}{\includegraphics[width=0.8\textwidth]{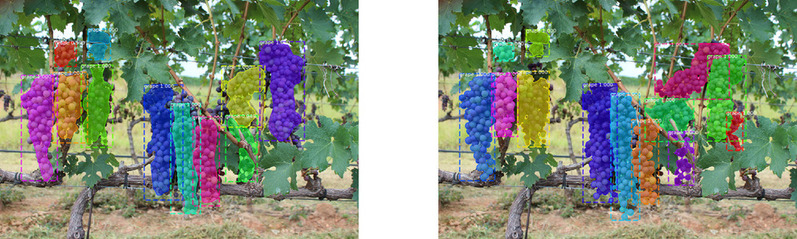}}\\
    \rotatebox{90}{\small \it Sauvignon Blanc} &
    \multicolumn{2}{c}{\includegraphics[width=0.8\textwidth]{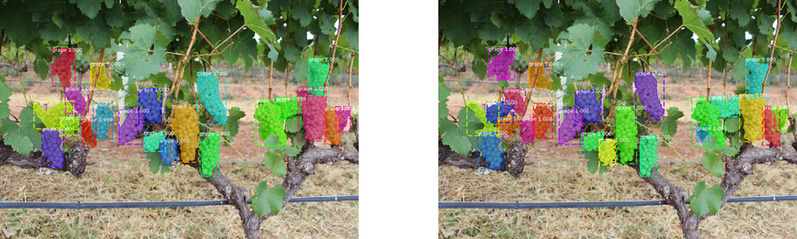}}\\
    \rotatebox{90}{\small \it Syrah} &
    \multicolumn{2}{c}{\includegraphics[width=0.8\textwidth]{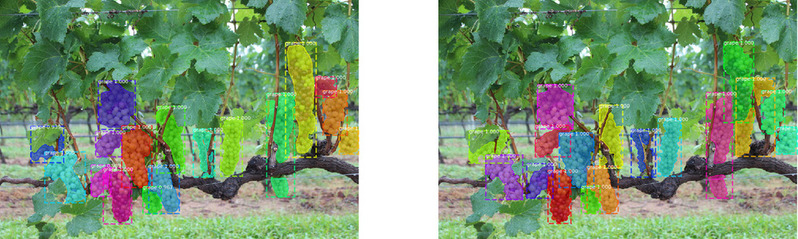}}\\
  \end{tabular}
  \caption{Some instance segmentation results produced by
    Mask~R-CNN, one example for each grape variety. (Left) Predictions by the network. (Right)
    Ground truth. Same color does not mean assignment between
    prediction and ground truth.} 
  \label{fig:InstanceSegmentation}
\end{figure}

Figure~\ref{fig:InstanceSegmentation} shows five examples of instance
segmentation results produced by the Mask~R-CNN. It illustrates the
network capability to learn shape, compactness and color
variability, and discriminate occluding foreground as branches and
trunks. Inter-variety color variation ({\it Chardonnay/Sauvignon
  Blanc vs. Cabernet/Syrah}) and intra-variety color variation ({\it
    Syrah} and {\it Cabernet} maturity) are properly modeled by the
  network, as well as shape, size and elongation ({\it Chardonnay vs. Cabernet},
  for example). The confidence level is also expressive: even
  considering that the confidence threshold is 0.9, most of the instances
  present levels equal or close to 1.0. Values lower than 0.99 can be
  observed in cases of severe occlusion, like the leftmost grape cluster
  in the {\it Syrah} example.

\begin{figure}
  \centering
  \begin{tabular}{lr}
    Predicted & Ground Truth\\
    \multicolumn{2}{c}{\includegraphics[width=\textwidth]{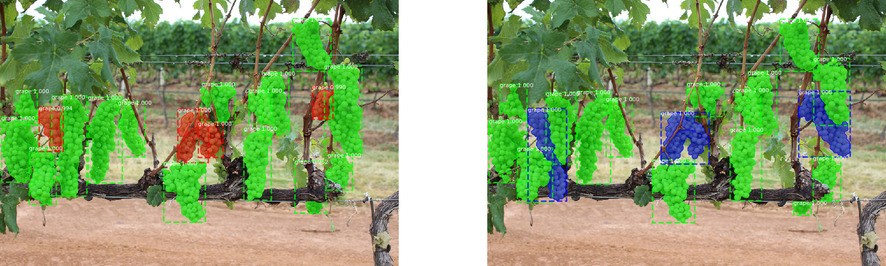}}\\
  \end{tabular}
  \caption{Divergence between predicted segmentation and the
    ground truth. (Left) Predictions by the network -- red clusters are
    false positives, green clusters are true positives. (Right) Ground truth
    -- blue clusters are false negatives. Disagreement in segmentation
    creates false negatives and false positives, despite correct
    detection of grape berries.} 
  \label{fig:SegmentationErrors}
\end{figure}

Grape cluster segmentation is challenging, even to the human
annotators: occlusions and the absence of 3-D input or on-site
annotation make the dataset error-prone regarding the correct
segmentation of large agglomerations of
clusters. {Figure~\ref{fig:SegmentationErrors} shows a 
case where segmentation divergence produces false negatives and false
positives in the evaluation, although the almost correct detection of
the grape berries. Note that for the clusters on the center in
Figure~\ref{fig:SegmentationErrors}, the prediction looks like a more
reasonable segmentation: two 
clusters are more plausible than one big and bifurcated cluster
proposed by the ground-truth, but we would need in-field inspection
(or a 3-D model) to check the correct detection. The other divergences
in the figure illustrate the difficulty of the segmentation task: the
clusters proposed by the prediction and the ground-truth look equally
plausible, and again only an in-field checking could provide the right
answer. Difficulties on successful cluster segmentation were also
reported by \citet{Nuske2014}. As noted before, segmentation divergence can also
deteriorate IoU. Fortunately, we will see further in this work than
data from a moving camera and 3-D association can relief such
occlusion issues by properly integrating images from multiple poses.}

$R_{\mathrm{inst}}$ and $P_{\mathrm{inst}}$ can suffer from erroneous segmentation, but what
about semantic segmentation? As can be seen in
Figure~\ref{fig:SegmentationErrors}, despite cluster segmentation errors,
at the berry level most of the grape pixels look properly
detected. To evaluate the detection of grape pixels, we use the
measures $R_{\mathrm{seman}}$ and $P_{\mathrm{seman}}$, recall and precision for the semantic
segmentation variation of the problem. Table~\ref{table:SemanticSegmentation} 
shows the overall result for
semantic segmentation on the entire masked set (last line), but also
the results found for each one of the 27 images. The table groups the
masked test set by the different varieties, allowing a comparison
across different grape types. The overall $F_1$ score for semantic
segmentation is 0.89 and no single variety has exhibited a remarkably
different score. {This is also an evidence that berries assignment to
individual clusters (cluster segmentation) is the main factor
affecting IoU.}

\begin{table}
  \caption{Semantic segmentation by Mask R-CNN. The first lines show
    evaluation for semantic segmentation (grape/background) for
    each image in the test set, stratified by variety for
    comparison. The last line shows the evaluation 
    for the entire test set (computed by accumulation of true
    positives, false positives and false negatives values).}
  \footnotesize
  \begin{tabular}{lrrr}
    \hline
      Image & $P_{\mathrm{seman}}$ & $R_{\mathrm{seman}}$ & $F_1$\\
      \hline
      CDY 2043 & 0.959 & 0.902 & 0.929\\
      CDY 2051 & 0.961 & 0.871 & 0.913\\
      CDY 2040 & 0.944 & 0.874 & 0.908\\
      CDY 2054 & 0.952 & 0.855 & 0.901\\
      CDY 2046 & 0.952 & 0.849 & 0.898\\
      CDY 2015 & 0.914 & 0.859 & 0.886\\
      \hline
      CFR 1638 & 0.928 & 0.885 & 0.906\\
      CFR 1641 & 0.899 & 0.873 & 0.886\\
      CFR 1639 & 0.930 & 0.841 & 0.883\\
      CFR 1643 & 0.918 & 0.835 & 0.875\\
      CFR 1666 & 0.951 & 0.807 & 0.873\\
      CFR 1651 & 0.906 & 0.808 & 0.854\\
      \hline
      CSV 20180427 144535647 & 0.937 & 0.898 & 0.917\\
      CSV 1877 & 0.928 & 0.879 & 0.903\\
      CSV 20180427 144507419 & 0.855 & 0.867 & 0.861\\
      CSV 1898 & 0.897 & 0.823 & 0.858\\
      CSV 20180427 144723166 & 0.850 & 0.848 & 0.849\\
      \hline
      SVB 20180427 151818928 & 0.949 & 0.890 & 0.919\\
      SVB 1954 & 0.912 & 0.915 & 0.913\\
      SVB 1944 & 0.900 & 0.922 & 0.911\\
      SVB 1935 & 0.926 & 0.856 & 0.889\\
      SVB 1972 & 0.895 & 0.860 & 0.877\\
      \hline
      SYH 2017-04-27 1318 & 0.943 & 0.866 & 0.903\\
      SYH 2017-04-27 1322 & 0.930 & 0.870 & 0.899\\
      SYH 2017-04-27 1239 & 0.921 & 0.867 & 0.893\\
      SYH 2017-04-27 1269 & 0.926 & 0.833 & 0.877\\
      SYH 2017-04-27 1304 & 0.908 & 0.746 & 0.819\\
      \hline
      All pixels in test set & {\bf 0.920} & {\bf 0.860} & {\bf
                                                           0.889}\\
    \hline
    \end{tabular}
    \label{table:SemanticSegmentation}
\end{table}

\begin{figure}
  \centering
  \begin{tabular}{ccccc}
    \space & Mask R-CNN & YOLOv2 & YOLOv3 & {Ground Truth}\\
    \rotatebox{90}{\small \it Chardonnay} &
    \includegraphics[width=0.23\textwidth]{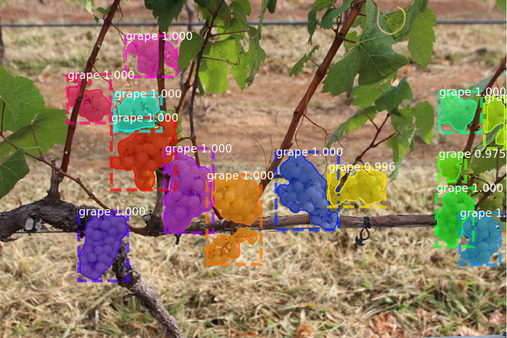} &
    \includegraphics[width=0.23\textwidth]{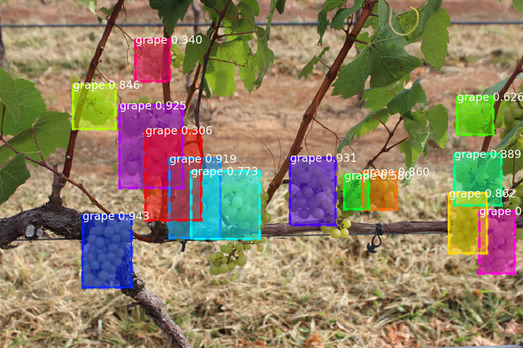} & 
    \includegraphics[width=0.23\textwidth]{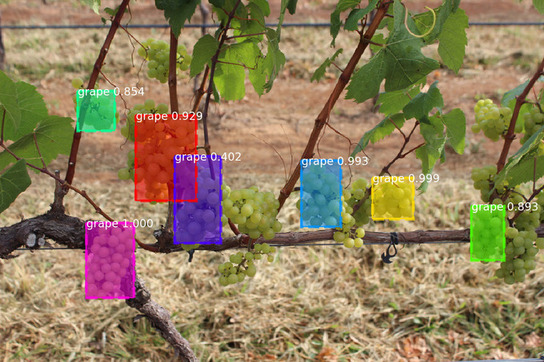}&
    \includegraphics[width=0.23\textwidth]{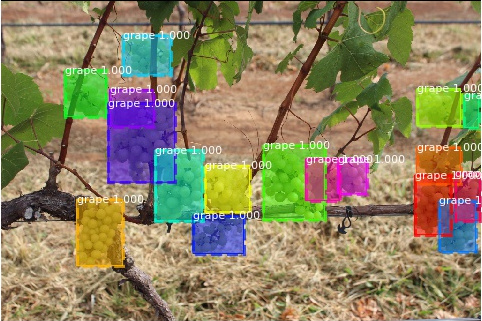}\\
    \rotatebox{90}{\small \it Cabernet Franc} &
    \includegraphics[width=0.23\textwidth]{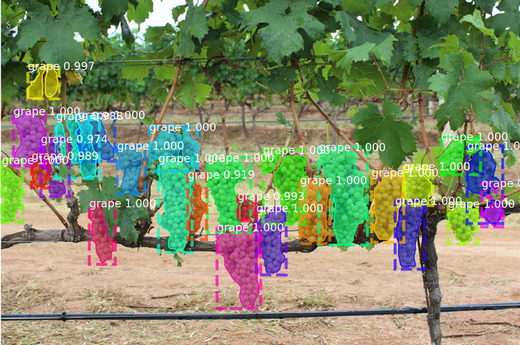} &
    \includegraphics[width=0.23\textwidth]{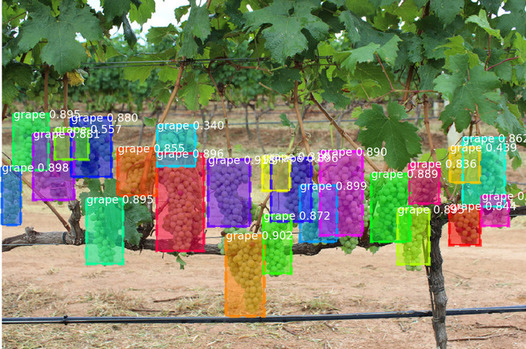} & 
    \includegraphics[width=0.23\textwidth]{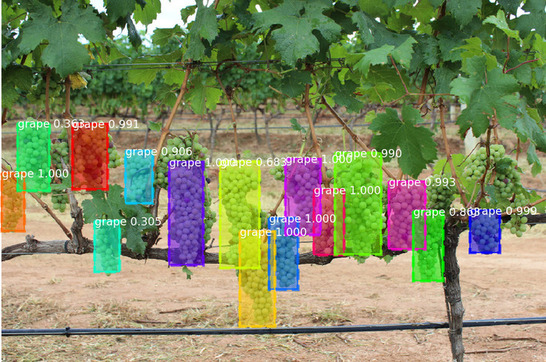}&
    \includegraphics[width=0.23\textwidth]{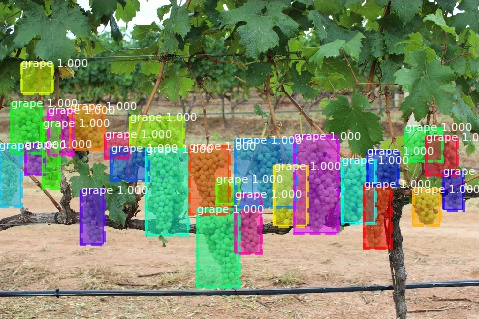}\\ 
    \rotatebox{90}{\small \it Cabernet Sauvignon} &
    \includegraphics[width=0.23\textwidth]{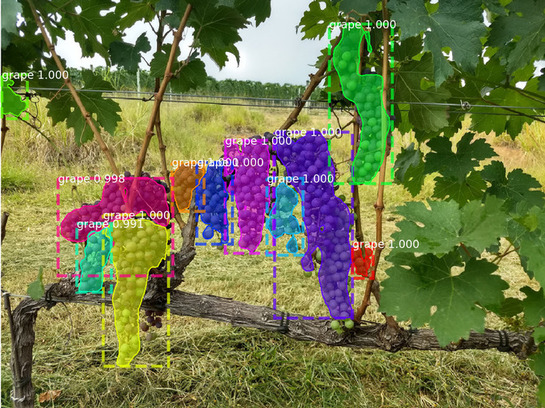} &
    \includegraphics[width=0.23\textwidth]{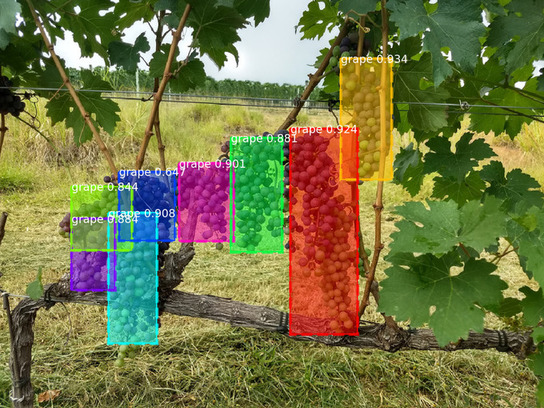} & 
    \includegraphics[width=0.23\textwidth]{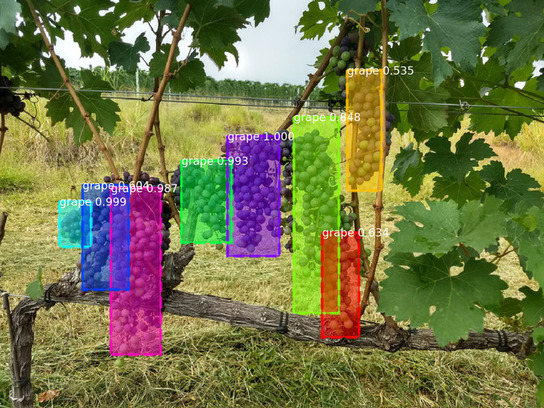} &
    \includegraphics[width=0.23\textwidth]{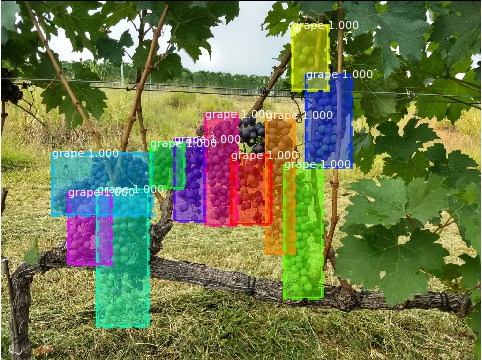}\\
    \rotatebox{90}{\small \it Sauvignon Blanc} &
    \includegraphics[width=0.23\textwidth]{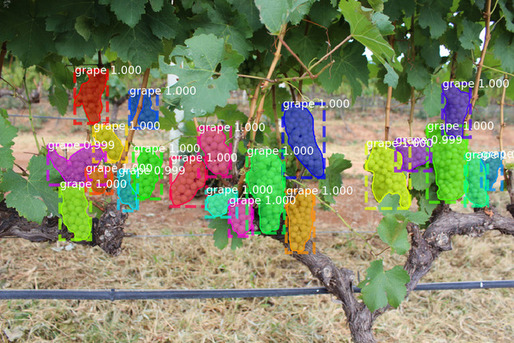} &
    \includegraphics[width=0.23\textwidth]{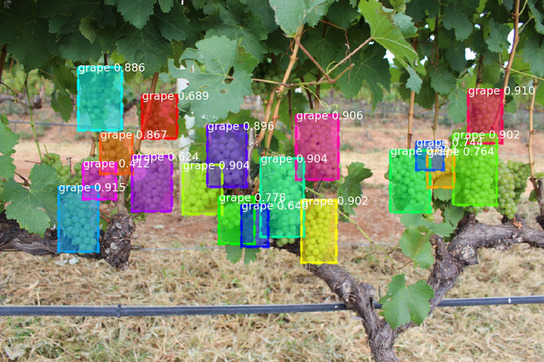} & 
    \includegraphics[width=0.23\textwidth]{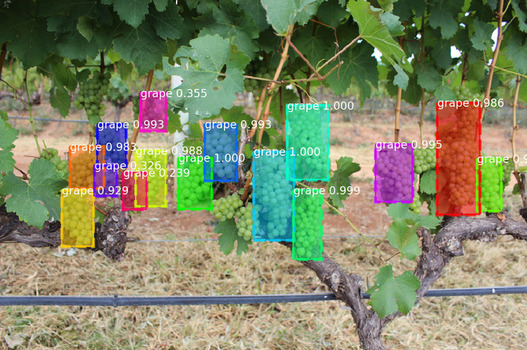} & 
    \includegraphics[width=0.23\textwidth]{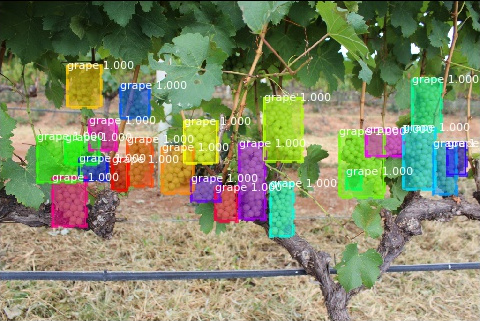}\\ 
    \rotatebox{90}{\small \it Syrah} &
    \includegraphics[width=0.23\textwidth]{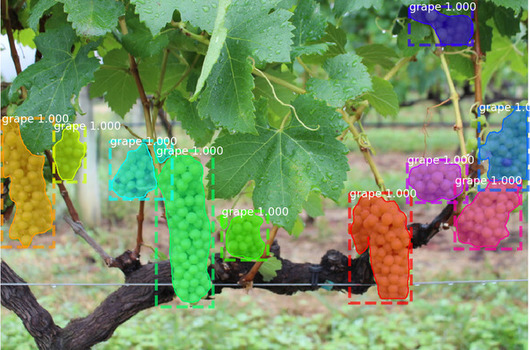} &
    \includegraphics[width=0.23\textwidth]{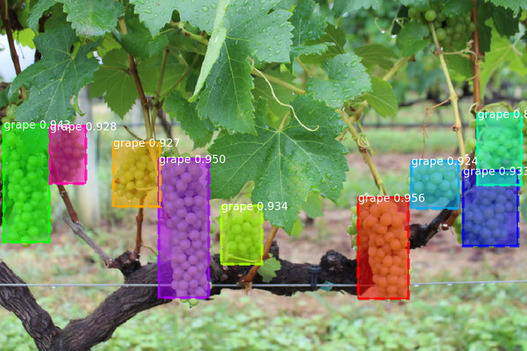} & 
    \includegraphics[width=0.23\textwidth]{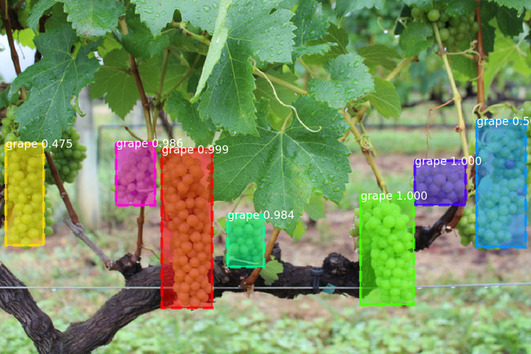} &
    \includegraphics[width=0.23\textwidth]{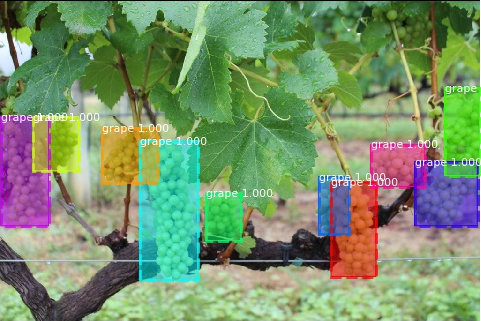}
  \end{tabular}
  \caption{Some object detection results produced by the three neural networks: Mask~R-CNN,
    YOLOv2 and YOLOv3, one example for each grape variety.
    Same color does not mean correspondence.} 
  \label{fig:ObjectDetection}
\end{figure}

Table~\ref{table:MaskRCNNAllBBox} presents the results for
object detection produced by the three networks, considering the entire test set of 837
clusters in 58 images. It is worth remembering that the models were trained using the
\emph{masked} training set, composed of 88 images (1,848 after
augmentation), but the results in Table~\ref{table:MaskRCNNAllBBox}
show the evaluation for the entire ``boxed'' test set (considering
intersection over union for the rectangular bounding boxes produced by
Mask~R-CNN). The recall values in the table show the YOLO networks
lose more clusters if compared to the Mask~R-CNN network, specially
YOLOv3. Figure~\ref{fig:ObjectDetection} shows some examples of
object detection for the three networks. In the figure we can see more
false negatives (lost clusters) in the YOLOv3's results. Also in
Table~\ref{table:MaskRCNNAllBBox}, as we require higher values of IoU
(what means better adjusted bounding boxes), YOLO networks show worse
results compared to Mask~R-CNN.        

\begin{table}
  \caption{Object detection for all test set of WGISD: Mask R-CNN, YOLOv2 and YOLOv3.}
  \footnotesize
  \begin{tabular}{rrrrrrrrrrrrrrr}
    \hline
      & \multicolumn{5}{c}{Mask~R-CNN} & \multicolumn{4}{c}{YOLOv2} & \multicolumn{4}{c}{YOLOv3}\\
      IoU & AP & $P_\mathrm{box}$ & $R_\mathrm{box}$ & $F_1$ & & AP & $P_\mathrm{box}$ & $R_\mathrm{box}$ & $F_1$
                                              & & AP & $P_\mathrm{box}$ & $R_\mathrm{box}$ & $F_1$\\
      \hline
      0.300 & 0.805 & 0.907 & 0.873 & 0.890 & & 0.675 & 0.893 & 0.728 & 0.802 & & 0.566 & 0.901 & 0.597 & 0.718\\
      0.400 & 0.777 & 0.891 & 0.858 & 0.874 & & 0.585 & 0.818 & 0.667 & 0.735 & & 0.494 & 0.829 & 0.550 & 0.661\\
      0.500 & 0.719 & 0.856 & 0.824 & 0.840 & & 0.478 & 0.726 & 0.591 & 0.652 & & 0.394 & 0.726 & 0.481 & 0.579\\
      0.600 & 0.611 & 0.788 & 0.759 & 0.773 & & 0.288 & 0.559 & 0.455 & 0.502 & & 0.261 & 0.587 & 0.389 & 0.468\\
      0.700 & 0.487 & 0.697 & 0.671 & 0.684 & & 0.139 & 0.390 & 0.318 & 0.350 & & 0.125 & 0.405 & 0.269 & 0.323\\
    0.800 & 0.276 & 0.521 & 0.502 & 0.511 & & 0.027 & 0.172 & 0.140 & 0.154 & & 0.036 & 0.205 & 0.136 & 0.164\\
    \hline
    \end{tabular}
  \label{table:MaskRCNNAllBBox}
\end{table}

\begin{figure}
  \begin{tabular}{c}
  Keyframe 135\\
  \includegraphics[width=0.9\textwidth]{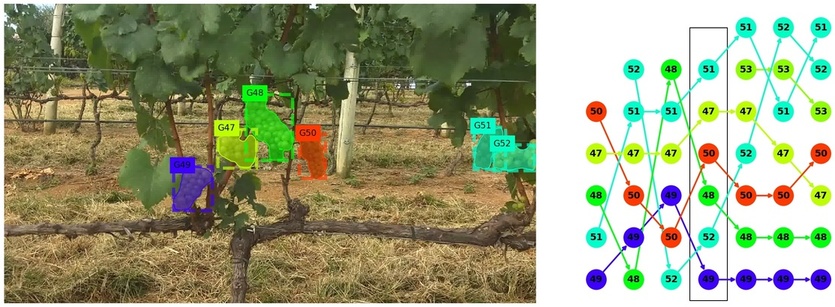}\\
  Keyframe 139\\
  \includegraphics[width=0.9\textwidth]{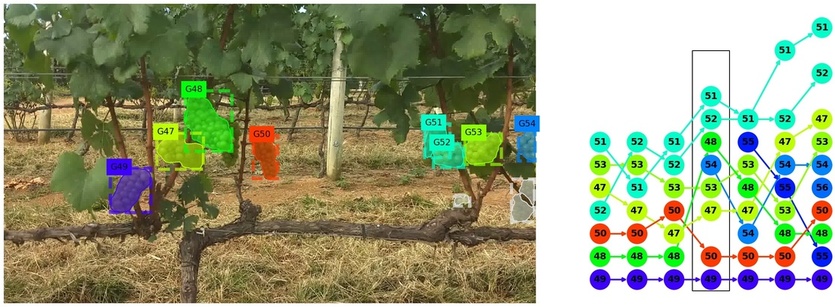}\\
  Keyframe 143\\
  \includegraphics[width=0.9\textwidth]{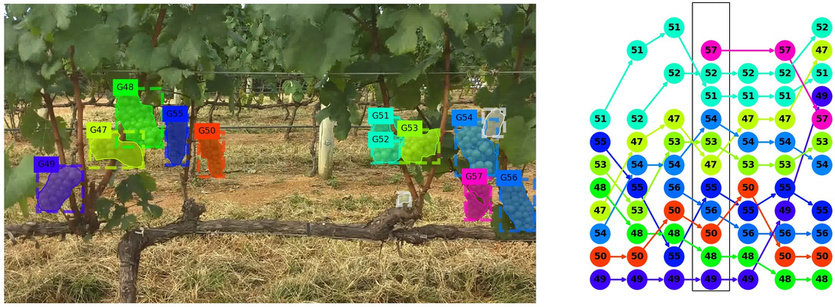}\\
  \end{tabular}
  \caption{Instance matching and tracking using 3-D assignment. (Left) Keyframes
  extracted from a video sequence by a 1080p camera. (Right) The graph-based tracking,
  similar to the one shown in Figure~\ref{fig:InstanceMatching}. Colors and numbers on the
  keyframes correspond to the colors and number in the graph. See the available 
  video for a more extensive demonstration.} 
  \label{fig:InstanceMatchingResults}
\end{figure}

To evaluate the spatial registration method and the potential of the
{entire methodology to address fruit counting, we employed a video
sequence captured on field. Multiple counting is avoided by the
tracking produced by the 3-D assignment: the number of individual
tracks should correspond to the number of observed clusters in the
video sequence, as seen previously in
Figure~\ref{fig:InstanceMatching}~(b). The video sequence was captured by a
smartphone camera} 
in full-HD ($1,920 \times 1,080$ pixels) while a service vehicle moved
along a row of vines. The keyframes of the MPEG video sequence were
extracted and the first 500 keyframes were employed in this
evaluation. Employing keyframes from the MPEG stream is useful
because (i) these frames present fewer compression artifacts than other frames
in the video sequence, (ii) the number of images (frames) is reduced,
and (iii) there is still sufficient overlap between frames to perform the
feature correspondence needed by structure-from-motion and to provide multiple
views for each grape cluster. Mask~R-CNN inference was performed for
each keyframe and the found mask stored. COLMAP was employed to create
a sparse 3-D model by SfM. Finally, the spatial registration proposed
on Section~\ref{sec:SpatialReg} was employed, matching the clusters
along the frame sequence
(Figure~\ref{fig:InstanceMatchingResults}). The results for the entire
frame sequence can be seen in an available
video\footnote{\url{https://youtu.be/1Hji3GS4mm4}. Note the video is
  edited to a 4 frames/second rate to allow the viewer follow the
  tracks more easily.}.     

\section{Discussion}
\label{sec:Discussion}

The Mask~R-CNN network presented superior results as compared to the YOLO networks. 
Considering IoU values equal or superior to 0.5, the advantage of Mask~R-CNN 
becomes more salient: even considering a 70\% IoU, the $F_1$ score is 
impressive. As a reference, \citet{Sa2016} reported 
0.828 and 0.848 $F_1$ scores for sweet peppers and rock melons respectively
at 0.4 IoU using Faster~R-CNN while \citet{Bargoti2017b}
reported a 0.90 $F_1$ for apples and mangoes considering a 0.2 IoU, also
employing Faster~R-CNN. However, readers should keep in mind that it is just
a reference, not a direct comparison or benchmark considering the different
crops and datasets.

The use of three different scales by YOLOv3 could not be an advantage
over YOLOv2 considering the almost constant distance between the camera and
the vineyard row. In the same way, Mask R-CNN's use of FPN could be reconsidered. 
Agronomical constraints could be explored: how big a group of berries 
should be to be considered a cluster? In other words, the operational and 
agronomical  context should be explored to define the scales of interest.
YOLOv3 employs multi-label classification, useful for problems presenting 
non-mutually exclusive object classes. However, considering our single class fruit detection
problems, this would not be an advantage of YOLOv3 compared to YOLOv2. Considering
that the YOLOv3 is deeper and, as consequence, prone to overfitting, it could need
more data to reach and surpass the results of YOLOv2, as observed in 
Table~\ref{table:MaskRCNNAllBBox}. {Although the better results
produced by Mask~R-CNN, it is important to note that rectangular bounding box
annotation for object detection is faster to be produced, what means
bigger datasets could be available to train networks like YOLOv2 and
YOLOv3 -- in this work, we have constrained YOLO training to the same
dataset available to Mask~R-CNN.}

\begin{figure}
  \centering
  \begin{tabular}{cc}
    \includegraphics[width=0.4\textwidth]{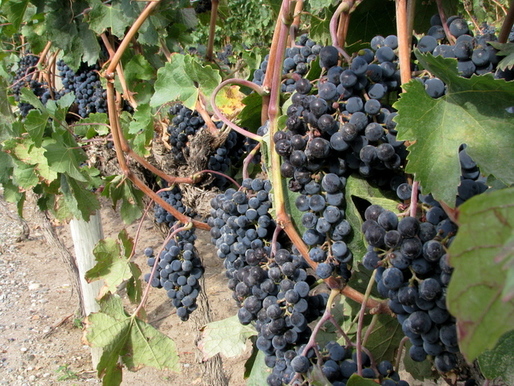} & \includegraphics[width=0.4\textwidth]{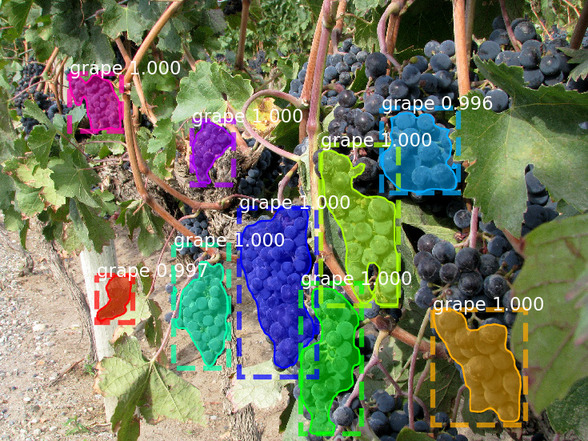}\\
    \multicolumn{2}{c}{{(a)}}\\
    \includegraphics[width=0.4\textwidth]{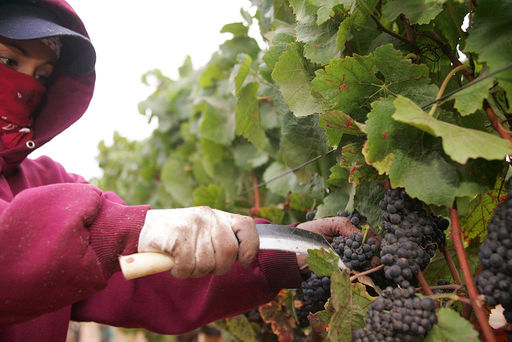} & \includegraphics[width=0.4\textwidth]{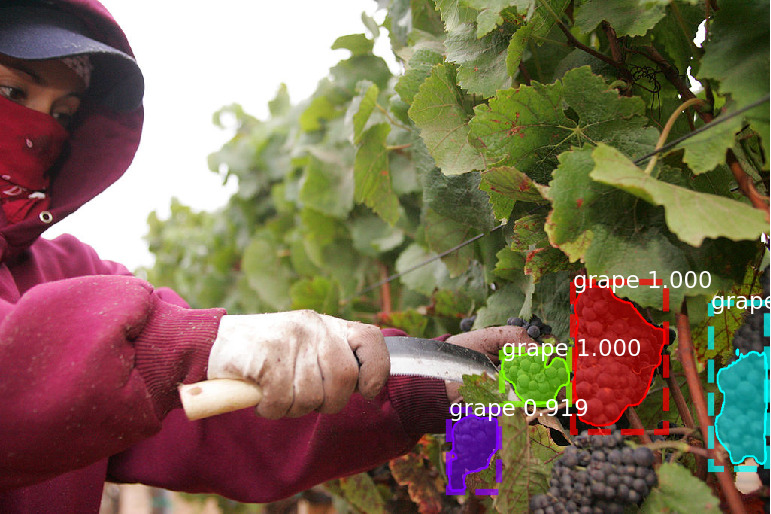}\\
    \multicolumn{2}{c}{{(b)}}\\
    \includegraphics[width=0.4\textwidth]{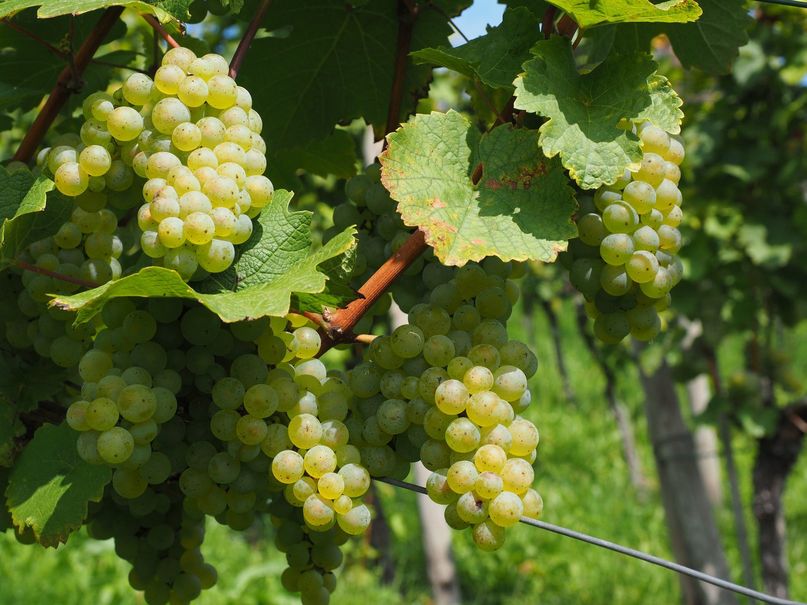} &\includegraphics[width=0.4\textwidth]{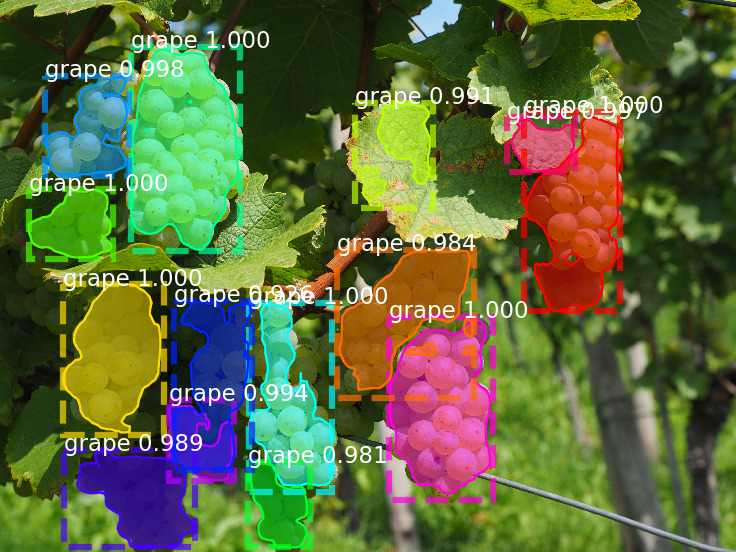}\\
    \multicolumn{2}{c}{{(c)}}\\
    \includegraphics[width=0.4\textwidth]{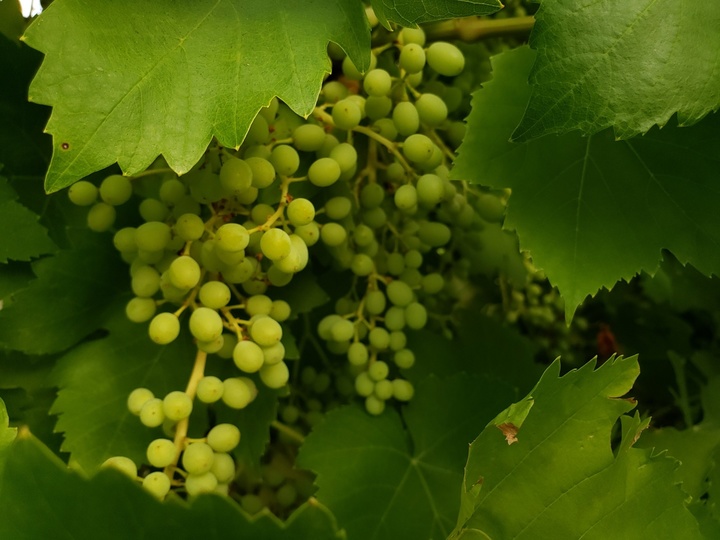} & \includegraphics[width=0.4\textwidth]{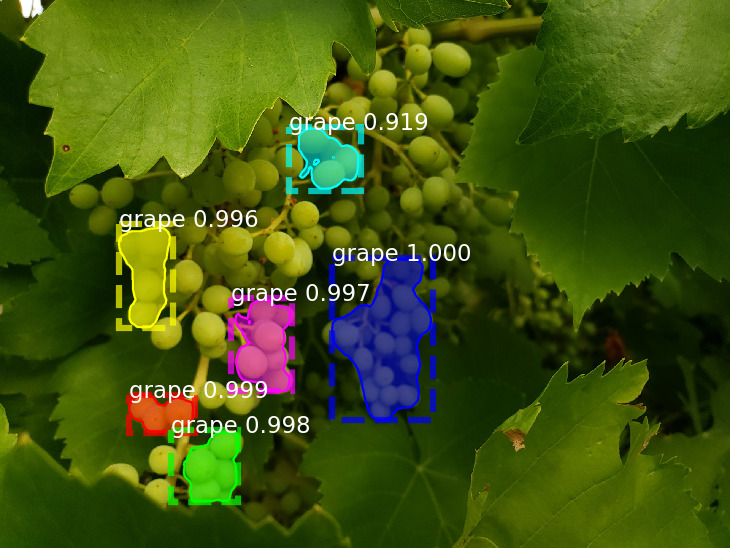}\\
    \multicolumn{2}{c}{{(d)}}
  \end{tabular}
  \caption{{Mask R-CNN generalization on novel
    scenarios without any
    tuning. (a)--(b) Examples presenting different camera poses. (c)
    Different pose and leaf texture. (d) Different developmental
    stage. These are Creative Commons-licensed images
     obtained from Internet (see text for references).}}
  \label{fig:XenoImages}
\end{figure}

{What could we be say about the \emph{generalization} of these models?
In other scenarios, like different developmental stages, crop
production systems or camera poses, they would be able to detect grape
clusters properly? 
Figure~\ref{fig:XenoImages} shows the results produced by the
Mask~R-CNN for images collected from the Internet\footnote{{The four
  images in Figure~\ref{fig:XenoImages} are public-domain or Creative
Commons-licensed pictures. Their authors are McKay Savage
(\url{https://www.flickr.com/photos/56796376@N00/4034779039}), Hahn
Family Wines
(\url{https://commons.wikimedia.org/wiki/File:Hand_harvesting_wine_grape.jpg}),
Peakpx (\url{http://www.peakpx.com/564748/green-grape-vine}) and Circe
Denyer
(\url{https://www.publicdomainpictures.net/en/view-image.php?image=298996}.}}). No
parameter 
tuning or other adaptions was
employed. Figures \ref{fig:XenoImages}~(a) and
\ref{fig:XenoImages}~(b) show  mature crops in
camera poses very different from the images in WGISD, the latter showing
a worker and a tool -- elements that are not present in the
dataset. The network was able to detect of a considerable
part of the instances. Figure~\ref{fig:XenoImages}~(c) shows an
example where false positives appeared in parts of textured leaves,
but most of the berries were detected and the predicted clusters looks
plausible. Finally, Figure~\ref{fig:XenoImages}~(d) shows clusters in
an earlier developmental stage. Apparently, the network is splitting
the clusters in regions of higher compactness, probably because the
training set in WGISD provides examples of more compact clusters. Such
qualitative results indicate the model present good generalization and
accurate results could be produced by tuning and transfer learning.}   

The presented CNN-based detectors can be integrated in larger systems
that, employing a data association strategy, will be able 
to integrate the detections and perform localized fruit
counting on site. {A moving camera, combined to fruit tracking, is a
powerful way to deal with occlusions due the integration of
different camera poses. Integrating video information  can also
relieve localized errors in fruit detection in a few frames.
As shown, an ordinary 1080p RGB camera can produce
input for accurate results, being an affordable approach to fruit
counting and orchard inspection.} Such vision systems can be easily
integrated in tractors, implements, service vehicles, robots and UAVs,
possibly employing high performance processing units (GPUs and TPUs)
with low energy consumption or even edge computing
\citep{Satyanarayanan2017}.

Notwithstanding, while our spatial integration is employing a
computational-intensive process such as structure-from-motion, other
implementations could use SLAM algorithms (simultaneous localization
and mapping), the real-time formulation 
of SfM. \citet{Liu2019} avoided the
computationally-intensive process of feature detection and matching in
SfM by employing the fruits' centers found by Faster R-CNN and Kalman Filter
tracking for inter-frame association. In other words, the fruits
became the \emph{landmarks} for the SfM procedure (implemented in
COLMAP). However, it is unclear what happens if \emph{no fruits} are
available in a segment of the video sequence. A fast SLAM algorithm such as
ORB-SLAM~\citep{Mur-Artal2017} or SVO~\citep{Forster2014}, not relying
on any specific landmark, could be a more robust alternative.

\section{Conclusion}
\label{sec:Conclusion}

Computer vision's current maturity level can produce impressive
and robust results in photogrammetry and perceptual tasks, even in 
challenging outdoor environments such as agricultural orchards. Combining
structure-from-motion (or its real-time version: SLAM) and
convolutional neural networks, advanced monitoring and robotics
applications can be developed for agriculture and livestock.

This work presents a methodology for grape detection, tracking and
counting in vineyards employing a single off-the-shelf 1080p
camera. We have reached $F_1$ scores superior to 0.9 for instance detection
in wine grapes, a challenging crop that presents enormous variability
in shape, size, color and compactness. We also showed that 3-D models
produced by structure-from-motion or SLAM can be employed to track
fruits, avoiding double counts and increasing tolerance to errors
in detection.  The same methodology could be used successfully for other
crops produced in trellis-like systems such as apples, peaches and
berries. Adaptions of the methodology can be  developed for fruits
grown in trees presenting bigger canopies, like citrus and mangoes --
yield could be estimated from regression from the visible fruit
counts.

Further research could consider more integration between the
photogrammetry and perception modules, looking for more
sophisticated \emph{scene understanding} systems able to robustly cope
with occlusions and other sources of errors.    

\section*{Acknowledgments}
\label{sec:Acks}

This work was supported by the Brazilian Agricultural Research Corporation
(Embrapa) under grant 01.14.09.001.05.04 and by CNPq PIBIC Program
(grants 161165/2017-6 and 125044/2018-6). S. Avila is partially funded by
Google LARA 2018 \& 2019, FAPESP (2017/16246-0) and FAEPEX (3125/17). We thank
Amy Tabb who provided helpful comments on the first version of the
manuscript. We also thank to Luís H. Bassoi and Luciano V. Koenigkan
for their generous support and help. Special thanks to Guaspari Winery
for allowing image data collection.  

\bibliographystyle{model5-names}\biboptions{authoryear}
\bibliography{references}

\begin{thebibliography}{57}
\expandafter\ifx\csname natexlab\endcsname\relax\def\natexlab#1{#1}\fi
\providecommand{\url}[1]{\texttt{#1}}
\providecommand{\href}[2]{#2}
\providecommand{\path}[1]{#1}
\providecommand{\DOIprefix}{doi:}
\providecommand{\ArXivprefix}{arXiv:}
\providecommand{\URLprefix}{URL: }
\providecommand{\Pubmedprefix}{pmid:}
\providecommand{\doi}[1]{\href{http://dx.doi.org/#1}{\path{#1}}}
\providecommand{\Pubmed}[1]{\href{pmid:#1}{\path{#1}}}
\providecommand{\bibinfo}[2]{#2}
\ifx\xfnm\relax \def\xfnm[#1]{\unskip,\space#1}\fi
\bibitem[{Acuna et~al.(2018)Acuna, Ling, Kar \& Fidler}]{Acuna2018}
\bibinfo{author}{Acuna, D.}, \bibinfo{author}{Ling, H.}, \bibinfo{author}{Kar,
  A.}, \& \bibinfo{author}{Fidler, S.} (\bibinfo{year}{2018}).
\newblock \bibinfo{title}{Efficient interactive annotation of segmentation
  datasets with {Polygon-RNN++}}.
\newblock In {\it \bibinfo{booktitle}{The IEEE Conference on Computer Vision
  and Pattern Recognition (CVPR)}\/}.
\bibitem[{Alahi et~al.(2012)Alahi, Ortiz \& Vandergheynst}]{Alahi2012}
\bibinfo{author}{Alahi, A.}, \bibinfo{author}{Ortiz, R.}, \&
  \bibinfo{author}{Vandergheynst, P.} (\bibinfo{year}{2012}).
\newblock \bibinfo{title}{{FREAK: Fast Retina Keypoint}}.
\newblock In {\it \bibinfo{booktitle}{2012 IEEE Conference on Computer Vision
  and Pattern Recognition}\/} (pp. \bibinfo{pages}{510--517}).
\newblock \bibinfo{publisher}{IEEE}.
\newblock \URLprefix \url{http://ieeexplore.ieee.org/document/6247715/}.
  \DOIprefix\doi{10.1109/CVPR.2012.6247715}.
\bibitem[{Barbedo(2019)}]{Barbedo2019}
\bibinfo{author}{Barbedo, J. G.~A.} (\bibinfo{year}{2019}).
\newblock \bibinfo{title}{{Plant disease identification from individual lesions
  and spots using deep learning}}.
\newblock {\it \bibinfo{journal}{Biosystems Engineering}\/},  {\it
  \bibinfo{volume}{180}\/}, \bibinfo{pages}{96--107}.
  \DOIprefix\doi{10.1016/J.BIOSYSTEMSENG.2019.02.002}.
\bibitem[{Bargoti \& Underwood(2017{\natexlab{a}})}]{Bargoti2017b}
\bibinfo{author}{Bargoti, S.}, \& \bibinfo{author}{Underwood, J.}
  (\bibinfo{year}{2017}{\natexlab{a}}).
\newblock \bibinfo{title}{{Deep fruit detection in orchards}}.
\newblock In {\it \bibinfo{booktitle}{2017 IEEE International Conference on
  Robotics and Automation (ICRA)}\/} (pp. \bibinfo{pages}{3626--3633}).
\newblock \bibinfo{publisher}{IEEE}.
\newblock \DOIprefix\doi{10.1109/ICRA.2017.7989417}.
  \href{http://arxiv.org/abs/arXiv:1610.03677v2}{\tt arXiv:arXiv:1610.03677v2}.
\bibitem[{Bargoti \& Underwood(2017{\natexlab{b}})}]{Bargoti2017a}
\bibinfo{author}{Bargoti, S.}, \& \bibinfo{author}{Underwood, J.~P.}
  (\bibinfo{year}{2017}{\natexlab{b}}).
\newblock \bibinfo{title}{{Image Segmentation for Fruit Detection and Yield
  Estimation in Apple Orchards}}.
\newblock {\it \bibinfo{journal}{Journal of Field Robotics}\/},  {\it
  \bibinfo{volume}{34}\/}, \bibinfo{pages}{1039--1060}.
  \DOIprefix\doi{10.1002/rob.21699}.
\bibitem[{Bengio et~al.(2013)Bengio, Courville \& Vincent}]{Bengio2013}
\bibinfo{author}{Bengio, Y.}, \bibinfo{author}{Courville, A.}, \&
  \bibinfo{author}{Vincent, P.} (\bibinfo{year}{2013}).
\newblock \bibinfo{title}{{Representation Learning: A Review and New
  Perspectives}}.
\newblock {\it \bibinfo{journal}{IEEE Transactions on Pattern Analysis and
  Machine Intelligence}\/},  {\it \bibinfo{volume}{35}\/},
  \bibinfo{pages}{1798--1828}. \DOIprefix\doi{10.1109/TPAMI.2013.50}.
\bibitem[{Chen et~al.(2017)Chen, Shivakumar, Dcunha, Das, Okon, Qu, Taylor \&
  Kumar}]{Chen2017}
\bibinfo{author}{Chen, S.~W.}, \bibinfo{author}{Shivakumar, S.~S.},
  \bibinfo{author}{Dcunha, S.}, \bibinfo{author}{Das, J.},
  \bibinfo{author}{Okon, E.}, \bibinfo{author}{Qu, C.},
  \bibinfo{author}{Taylor, C.~J.}, \& \bibinfo{author}{Kumar, V.}
  (\bibinfo{year}{2017}).
\newblock \bibinfo{title}{{Counting Apples and Oranges With Deep Learning: A
  Data-Driven Approach}}.
\newblock {\it \bibinfo{journal}{IEEE Robotics and Automation Letters}\/},
  {\it \bibinfo{volume}{2}\/}, \bibinfo{pages}{781--788}.
  \DOIprefix\doi{10.1109/LRA.2017.2651944}.
\bibitem[{Chollet(2017)}]{Chollet2017}
\bibinfo{author}{Chollet, F.} (\bibinfo{year}{2017}).
\newblock {\it \bibinfo{title}{Deep learning with Python}\/}.
\newblock \bibinfo{publisher}{Manning Publications Company}.
\bibitem[{Deng et~al.(2009)Deng, Dong, Socher, Li, Li \&
  Fei-Fei}]{Deng2009ImageNet}
\bibinfo{author}{Deng, J.}, \bibinfo{author}{Dong, W.},
  \bibinfo{author}{Socher, R.}, \bibinfo{author}{Li, L.-J.},
  \bibinfo{author}{Li, K.}, \& \bibinfo{author}{Fei-Fei, L.}
  (\bibinfo{year}{2009}).
\newblock \bibinfo{title}{Imagenet: A large-scale hierarchical image database}.
\newblock In {\it \bibinfo{booktitle}{2009 IEEE Conference on Computer Vision
  and Pattern Recognition}\/} (pp. \bibinfo{pages}{248--255}).
\newblock \bibinfo{organization}{Ieee}.
\bibitem[{Duckett et~al.(2018)Duckett, Pearson, Blackmore \&
  Grieve}]{UKRAS2018}
\bibinfo{author}{Duckett, T.}, \bibinfo{author}{Pearson, S.},
  \bibinfo{author}{Blackmore, S.}, \& \bibinfo{author}{Grieve, B.}
  (\bibinfo{year}{2018}).
\newblock \bibinfo{title}{Agricultural robotics: The future of robotic
  agriculture}.
\newblock {\it \bibinfo{journal}{CoRR}\/},  {\it
  \bibinfo{volume}{abs/1806.06762}\/}. \URLprefix
  \url{http://arxiv.org/abs/1806.06762}.
  \href{http://arxiv.org/abs/1806.06762}{\tt arXiv:1806.06762}.
\bibitem[{Dunn \& Martin(2004)}]{Dunn2004}
\bibinfo{author}{Dunn, G.~M.}, \& \bibinfo{author}{Martin, S.~R.}
  (\bibinfo{year}{2004}).
\newblock \bibinfo{title}{{Yield prediction from digital image analysis: A
  technique with potential for vineyard assessments prior to harvest}}.
\newblock {\it \bibinfo{journal}{Australian Journal of Grape and Wine
  Research}\/},  {\it \bibinfo{volume}{10}\/}, \bibinfo{pages}{196--198}.
  \DOIprefix\doi{10.1111/j.1755-0238.2004.tb00022.x}.
\bibitem[{Dutta et~al.(2016)Dutta, Gupta \& Zissermann}]{Dutta2016}
\bibinfo{author}{Dutta, A.}, \bibinfo{author}{Gupta, A.}, \&
  \bibinfo{author}{Zissermann, A.} (\bibinfo{year}{2016}).
\newblock \bibinfo{title}{{VGG} image annotator ({VIA})}.
\newblock
  \bibinfo{howpublished}{\url{http://www.robots.ox.ac.uk/~vgg/software/via/}}.
\newblock \bibinfo{note}{Version: 2.0.6, Accessed: April 23, 2019}.
\bibitem[{Everingham et~al.(2010)Everingham, Van~Gool, Williams, Winn \&
  Zisserman}]{PascalVOC}
\bibinfo{author}{Everingham, M.}, \bibinfo{author}{Van~Gool, L.},
  \bibinfo{author}{Williams, C.~K.}, \bibinfo{author}{Winn, J.}, \&
  \bibinfo{author}{Zisserman, A.} (\bibinfo{year}{2010}).
\newblock \bibinfo{title}{The {Pascal} {V}isual {O}bject {C}lasses ({VOC})
  {C}hallenge}.
\newblock {\it \bibinfo{journal}{International Journal of Computer Vision}\/},
  {\it \bibinfo{volume}{88}\/}, \bibinfo{pages}{303--338}.
\bibitem[{Forster et~al.(2014)Forster, Pizzoli \& Scaramuzza}]{Forster2014}
\bibinfo{author}{Forster, C.}, \bibinfo{author}{Pizzoli, M.}, \&
  \bibinfo{author}{Scaramuzza, D.} (\bibinfo{year}{2014}).
\newblock \bibinfo{title}{Svo: Fast semi-direct monocular visual odometry}.
\newblock In {\it \bibinfo{booktitle}{2014 IEEE international conference on
  robotics and automation (ICRA)}\/} (pp. \bibinfo{pages}{15--22}).
\newblock \bibinfo{organization}{IEEE}.
\bibitem[{Gebru et~al.(2018)Gebru, Morgenstern, Vecchione, Vaughan, Wallach,
  III \& Crawford}]{Gebru2018}
\bibinfo{author}{Gebru, T.}, \bibinfo{author}{Morgenstern, J.},
  \bibinfo{author}{Vecchione, B.}, \bibinfo{author}{Vaughan, J.~W.},
  \bibinfo{author}{Wallach, H.~M.}, \bibinfo{author}{III, H.~D.}, \&
  \bibinfo{author}{Crawford, K.} (\bibinfo{year}{2018}).
\newblock \bibinfo{title}{Datasheets for datasets}.
\newblock {\it \bibinfo{journal}{CoRR}\/},  {\it
  \bibinfo{volume}{abs/1803.09010}\/}. \URLprefix
  \url{http://arxiv.org/abs/1803.09010}.
  \href{http://arxiv.org/abs/1803.09010}{\tt arXiv:1803.09010}.
\bibitem[{Girshick(2015)}]{Girshick2015}
\bibinfo{author}{Girshick, R.} (\bibinfo{year}{2015}).
\newblock \bibinfo{title}{Fast {R-CNN}}.
\newblock In {\it \bibinfo{booktitle}{Proceedings of the IEEE international
  conference on computer vision}\/} (pp. \bibinfo{pages}{1440--1448}).
\bibitem[{Gongal et~al.(2015)Gongal, Amatya, Karkee, Zhang \&
  Lewis}]{Gongal2015}
\bibinfo{author}{Gongal, A.}, \bibinfo{author}{Amatya, S.},
  \bibinfo{author}{Karkee, M.}, \bibinfo{author}{Zhang, Q.}, \&
  \bibinfo{author}{Lewis, K.} (\bibinfo{year}{2015}).
\newblock \bibinfo{title}{{Sensors and systems for fruit detection and
  localization: A review}}.
\newblock {\it \bibinfo{journal}{Computers and Electronics in Agriculture}\/},
  {\it \bibinfo{volume}{116}\/}, \bibinfo{pages}{8--19}.
  \DOIprefix\doi{10.1016/j.compag.2015.05.021}.
\bibitem[{Goodfellow et~al.(2016)Goodfellow, Bengio \&
  Courville}]{Goodfellow-et-al-2016}
\bibinfo{author}{Goodfellow, I.}, \bibinfo{author}{Bengio, Y.}, \&
  \bibinfo{author}{Courville, A.} (\bibinfo{year}{2016}).
\newblock {\it \bibinfo{title}{Deep Learning}\/}.
\newblock \bibinfo{publisher}{MIT Press}.
\newblock \bibinfo{note}{\url{http://www.deeplearningbook.org}}.
\bibitem[{Hartley \& Zisserman(2003)}]{HZ2003}
\bibinfo{author}{Hartley, R.}, \& \bibinfo{author}{Zisserman, A.}
  (\bibinfo{year}{2003}).
\newblock {\it \bibinfo{title}{Multiple View Geometry in Computer Vision}\/}.
\newblock (\bibinfo{edition}{2nd} ed.).
\newblock \bibinfo{address}{New York, NY, USA}: \bibinfo{publisher}{Cambridge
  University Press}.
\bibitem[{He et~al.(2017)He, Gkioxari, Dollar \& Girshick}]{He_2017_ICCV}
\bibinfo{author}{He, K.}, \bibinfo{author}{Gkioxari, G.},
  \bibinfo{author}{Dollar, P.}, \& \bibinfo{author}{Girshick, R.}
  (\bibinfo{year}{2017}).
\newblock \bibinfo{title}{Mask {R-CNN}}.
\newblock In {\it \bibinfo{booktitle}{The IEEE International Conference on
  Computer Vision (ICCV)}\/}.
\bibitem[{He et~al.(2016)He, Zhang, Ren \& Sun}]{He2016}
\bibinfo{author}{He, K.}, \bibinfo{author}{Zhang, X.}, \bibinfo{author}{Ren,
  S.}, \& \bibinfo{author}{Sun, J.} (\bibinfo{year}{2016}).
\newblock \bibinfo{title}{{Deep Residual Learning for Image Recognition}}.
\newblock In {\it \bibinfo{booktitle}{2016 IEEE Conference on Computer Vision
  and Pattern Recognition (CVPR)}\/} (pp. \bibinfo{pages}{770--778}).
\newblock \bibinfo{publisher}{IEEE}.
\newblock \DOIprefix\doi{10.1109/CVPR.2016.90}.
\bibitem[{Huang et~al.(2017)Huang, Rathod, Sun, Zhu, Korattikara, Fathi,
  Fischer, Wojna, Song, Guadarrama \& Murphy}]{Huang2017}
\bibinfo{author}{Huang, J.}, \bibinfo{author}{Rathod, V.},
  \bibinfo{author}{Sun, C.}, \bibinfo{author}{Zhu, M.},
  \bibinfo{author}{Korattikara, A.}, \bibinfo{author}{Fathi, A.},
  \bibinfo{author}{Fischer, I.}, \bibinfo{author}{Wojna, Z.},
  \bibinfo{author}{Song, Y.}, \bibinfo{author}{Guadarrama, S.}, \&
  \bibinfo{author}{Murphy, K.} (\bibinfo{year}{2017}).
\newblock \bibinfo{title}{Speed/accuracy trade-offs for modern convolutional
  object detectors}.
\newblock In {\it \bibinfo{booktitle}{The IEEE Conference on Computer Vision
  and Pattern Recognition (CVPR)}\/}.
\bibitem[{Jung(2019)}]{imgaug}
\bibinfo{author}{Jung, A.} (\bibinfo{year}{2019}).
\newblock \bibinfo{title}{{imgaug} documentation}.
\newblock \bibinfo{howpublished}{\url{https://imgaug.readthedocs.io}}.
\newblock \bibinfo{note}{Revision: cce07845, Accessed: June 23, 2019}.
\bibitem[{Kamilaris \& Prenafeta-Bold{\'{u}}(2018)}]{Kamilaris2018}
\bibinfo{author}{Kamilaris, A.}, \& \bibinfo{author}{Prenafeta-Bold{\'{u}},
  F.~X.} (\bibinfo{year}{2018}).
\newblock \bibinfo{title}{{Deep learning in agriculture: A survey}}.
\newblock {\it \bibinfo{journal}{Computers and Electronics in Agriculture}\/},
  {\it \bibinfo{volume}{147}\/}, \bibinfo{pages}{70--90}.
  \DOIprefix\doi{10.1016/J.COMPAG.2018.02.016}.
\bibitem[{Kicherer et~al.(2017)Kicherer, Herzog, Bendel, Klück, Backhaus,
  Wieland, Rose, Klingbeil, Läbe, Hohl, Petry, Kuhlmann, Seiffert \&
  Töpfer}]{Kicherer2017}
\bibinfo{author}{Kicherer, A.}, \bibinfo{author}{Herzog, K.},
  \bibinfo{author}{Bendel, N.}, \bibinfo{author}{Klück, H.-C.},
  \bibinfo{author}{Backhaus, A.}, \bibinfo{author}{Wieland, M.},
  \bibinfo{author}{Rose, J.~C.}, \bibinfo{author}{Klingbeil, L.},
  \bibinfo{author}{Läbe, T.}, \bibinfo{author}{Hohl, C.},
  \bibinfo{author}{Petry, W.}, \bibinfo{author}{Kuhlmann, H.},
  \bibinfo{author}{Seiffert, U.}, \& \bibinfo{author}{Töpfer, R.}
  (\bibinfo{year}{2017}).
\newblock \bibinfo{title}{Phenoliner: A new field phenotyping platform for
  grapevine research}.
\newblock {\it \bibinfo{journal}{Sensors}\/},  {\it \bibinfo{volume}{17}\/}.
  \DOIprefix\doi{10.3390/s17071625}.
\bibitem[{Kirkpatrick(2019)}]{Kirkpatrick2019}
\bibinfo{author}{Kirkpatrick, K.} (\bibinfo{year}{2019}).
\newblock \bibinfo{title}{{Technologizing agriculture}}.
\newblock {\it \bibinfo{journal}{Communications of the ACM}\/},  {\it
  \bibinfo{volume}{62}\/}, \bibinfo{pages}{14--16}.
  \DOIprefix\doi{10.1145/3297805}.
\bibitem[{Krizhevsky et~al.(2012)Krizhevsky, Sutskever \&
  Hinton}]{NIPS2012_4824}
\bibinfo{author}{Krizhevsky, A.}, \bibinfo{author}{Sutskever, I.}, \&
  \bibinfo{author}{Hinton, G.~E.} (\bibinfo{year}{2012}).
\newblock \bibinfo{title}{{ImageNet Classification with Deep Convolutional
  Neural Networks}}.
\newblock In \bibinfo{editor}{F.~Pereira}, \bibinfo{editor}{C.~J.~C. Burges},
  \bibinfo{editor}{L.~Bottou}, \& \bibinfo{editor}{K.~Q. Weinberger} (Eds.),
  {\it \bibinfo{booktitle}{Advances in Neural Information Processing Systems
  25}\/} (pp. \bibinfo{pages}{1097--1105}).
\newblock \bibinfo{publisher}{Curran Associates, Inc.}
\bibitem[{LeCun et~al.(2015)LeCun, Bengio \& Hinton}]{LeCun2015}
\bibinfo{author}{LeCun, Y.}, \bibinfo{author}{Bengio, Y.}, \&
  \bibinfo{author}{Hinton, G.} (\bibinfo{year}{2015}).
\newblock \bibinfo{title}{{Deep learning}}.
\newblock {\it \bibinfo{journal}{Nature}\/},  {\it \bibinfo{volume}{521}\/},
  \bibinfo{pages}{436--444}. \DOIprefix\doi{10.1038/nature14539}.
\bibitem[{Lin et~al.(2017)Lin, Dollar, Girshick, He, Hariharan \&
  Belongie}]{Lin_2017_CVPR}
\bibinfo{author}{Lin, T.-Y.}, \bibinfo{author}{Dollar, P.},
  \bibinfo{author}{Girshick, R.}, \bibinfo{author}{He, K.},
  \bibinfo{author}{Hariharan, B.}, \& \bibinfo{author}{Belongie, S.}
  (\bibinfo{year}{2017}).
\newblock \bibinfo{title}{Feature pyramid networks for object detection}.
\newblock In {\it \bibinfo{booktitle}{The IEEE Conference on Computer Vision
  and Pattern Recognition (CVPR)}\/}.
\bibitem[{Lin et~al.(2014)Lin, Maire, Belongie, Hays, Perona, Ramanan,
  Doll{\'{a}}r \& Zitnick}]{MSCOCO}
\bibinfo{author}{Lin, T.-Y.}, \bibinfo{author}{Maire, M.},
  \bibinfo{author}{Belongie, S.}, \bibinfo{author}{Hays, J.},
  \bibinfo{author}{Perona, P.}, \bibinfo{author}{Ramanan, D.},
  \bibinfo{author}{Doll{\'{a}}r, P.}, \& \bibinfo{author}{Zitnick, C.~L.}
  (\bibinfo{year}{2014}).
\newblock \bibinfo{title}{{Microsoft COCO: Common Objects in Context}}.
\newblock In \bibinfo{editor}{D.~Fleet}, \bibinfo{editor}{T.~Pajdla},
  \bibinfo{editor}{B.~Schiele}, \& \bibinfo{editor}{T.~Tuytelaars} (Eds.), {\it
  \bibinfo{booktitle}{Computer Vision -- ECCV 2014}\/} (pp.
  \bibinfo{pages}{740--755}).
\newblock \bibinfo{publisher}{Springer International Publishing}.
\bibitem[{Liu et~al.(2018)Liu, Ouyang, Wang, Fieguth, Chen, Liu \&
  Pietik{\"{a}}inen}]{Liu2018}
\bibinfo{author}{Liu, L.}, \bibinfo{author}{Ouyang, W.}, \bibinfo{author}{Wang,
  X.}, \bibinfo{author}{Fieguth, P.}, \bibinfo{author}{Chen, J.},
  \bibinfo{author}{Liu, X.}, \& \bibinfo{author}{Pietik{\"{a}}inen, M.}
  (\bibinfo{year}{2018}).
\newblock \bibinfo{title}{{Deep Learning for Generic Object Detection: A
  Survey}}, .
\newblock \URLprefix \url{http://arxiv.org/abs/1809.02165}.
  \href{http://arxiv.org/abs/1809.02165}{\tt arXiv:1809.02165}.
\bibitem[{Liu et~al.(2016)Liu, Anguelov, Erhan, Szegedy, Reed, Fu \&
  Berg}]{Liu2016}
\bibinfo{author}{Liu, W.}, \bibinfo{author}{Anguelov, D.},
  \bibinfo{author}{Erhan, D.}, \bibinfo{author}{Szegedy, C.},
  \bibinfo{author}{Reed, S.}, \bibinfo{author}{Fu, C.-Y.}, \&
  \bibinfo{author}{Berg, A.~C.} (\bibinfo{year}{2016}).
\newblock \bibinfo{title}{Ssd: Single shot multibox detector}.
\newblock In {\it \bibinfo{booktitle}{European conference on computer
  vision}\/} (pp. \bibinfo{pages}{21--37}).
\newblock \bibinfo{organization}{Springer}.
\bibitem[{Liu et~al.(2019)Liu, Chen, Liu, Shivakumar, Das, Taylor, Underwood \&
  Kumar}]{Liu2019}
\bibinfo{author}{Liu, X.}, \bibinfo{author}{Chen, S.~W.}, \bibinfo{author}{Liu,
  C.}, \bibinfo{author}{Shivakumar, S.~S.}, \bibinfo{author}{Das, J.},
  \bibinfo{author}{Taylor, C.~J.}, \bibinfo{author}{Underwood, J.}, \&
  \bibinfo{author}{Kumar, V.} (\bibinfo{year}{2019}).
\newblock \bibinfo{title}{{Monocular Camera Based Fruit Counting and Mapping
  With Semantic Data Association}}.
\newblock {\it \bibinfo{journal}{IEEE Robotics and Automation Letters}\/},
  {\it \bibinfo{volume}{4}\/}, \bibinfo{pages}{2296--2303}.
  \DOIprefix\doi{10.1109/LRA.2019.2901987}.
\bibitem[{Lowe(2004)}]{LoweSIFT2004}
\bibinfo{author}{Lowe, D.~G.} (\bibinfo{year}{2004}).
\newblock \bibinfo{title}{{Distinctive Image Features from Scale-Invariant
  Keypoints}}.
\newblock {\it \bibinfo{journal}{International Journal of Computer Vision}\/},
  {\it \bibinfo{volume}{60}\/}, \bibinfo{pages}{91--110}.
  \DOIprefix\doi{10.1023/B:VISI.0000029664.99615.94}.
\bibitem[{Loy \& Zelinsky(2003)}]{Loy2003}
\bibinfo{author}{Loy, G.}, \& \bibinfo{author}{Zelinsky, A.}
  (\bibinfo{year}{2003}).
\newblock \bibinfo{title}{{Fast radial symmetry for detecting points of
  interest}}.
\newblock {\it \bibinfo{journal}{IEEE Transactions on Pattern Analysis and
  Machine Intelligence}\/},  {\it \bibinfo{volume}{25}\/},
  \bibinfo{pages}{959--973}. \DOIprefix\doi{10.1109/TPAMI.2003.1217601}.
\bibitem[{{Matterport, Inc}(2018)}]{Matterport}
\bibinfo{author}{{Matterport, Inc}} (\bibinfo{year}{2018}).
\newblock \bibinfo{title}{{{Mask R-CNN for Object Detection and
  Segmentation}}}.
\newblock
  \bibinfo{howpublished}{\url{https://github.com/matterport/Mask_RCNN}}.
\newblock \bibinfo{note}{Commit: 4f440de, Accessed: June 23, 2019.}
\bibitem[{Mur-Artal \& Tardos(2017)}]{Mur-Artal2017}
\bibinfo{author}{Mur-Artal, R.}, \& \bibinfo{author}{Tardos, J.~D.}
  (\bibinfo{year}{2017}).
\newblock \bibinfo{title}{{ORB-SLAM2: An Open-Source SLAM System for Monocular,
  Stereo, and RGB-D Cameras}}.
\newblock {\it \bibinfo{journal}{IEEE Transactions on Robotics}\/},  {\it
  \bibinfo{volume}{33}\/}, \bibinfo{pages}{1255--1262}.
  \DOIprefix\doi{10.1109/TRO.2017.2705103}.
  \href{http://arxiv.org/abs/1610.06475}{\tt arXiv:1610.06475}.
\bibitem[{Noma et~al.(2012)Noma, Graciano, Cesar, Consularo \&
  Bloch}]{Noma2012}
\bibinfo{author}{Noma, A.}, \bibinfo{author}{Graciano, A.~B.},
  \bibinfo{author}{Cesar, R.~M.}, \bibinfo{author}{Consularo, L.~A.}, \&
  \bibinfo{author}{Bloch, I.} (\bibinfo{year}{2012}).
\newblock \bibinfo{title}{{Interactive image segmentation by matching
  attributed relational graphs}}.
\newblock {\it \bibinfo{journal}{Pattern Recognition}\/},  {\it
  \bibinfo{volume}{45}\/}, \bibinfo{pages}{1159--1179}.
  \DOIprefix\doi{10.1016/j.patcog.2011.08.017}.
\bibitem[{Nuske et~al.(2011)Nuske, Achar, Bates, Narasimhan \&
  Singh}]{Nuske2011}
\bibinfo{author}{Nuske, S.}, \bibinfo{author}{Achar, S.},
  \bibinfo{author}{Bates, T.}, \bibinfo{author}{Narasimhan, S.}, \&
  \bibinfo{author}{Singh, S.} (\bibinfo{year}{2011}).
\newblock \bibinfo{title}{{Yield estimation in vineyards by visual grape
  detection}}.
\newblock In {\it \bibinfo{booktitle}{IEEE International Conference on
  Intelligent Robots and Systems}\/} (pp. \bibinfo{pages}{2352--2358}).
\newblock \DOIprefix\doi{10.1109/IROS.2011.6048830}.
\bibitem[{Nuske et~al.(2014)Nuske, Wilshusen, Achar, Yoder \&
  Singh}]{Nuske2014}
\bibinfo{author}{Nuske, S.}, \bibinfo{author}{Wilshusen, K.},
  \bibinfo{author}{Achar, S.}, \bibinfo{author}{Yoder, L.}, \&
  \bibinfo{author}{Singh, S.} (\bibinfo{year}{2014}).
\newblock \bibinfo{title}{{Automated visual yield estimation in vineyards}}.
\newblock {\it \bibinfo{journal}{Journal of Field Robotics}\/},  {\it
  \bibinfo{volume}{31}\/}, \bibinfo{pages}{837--860}.
  \DOIprefix\doi{10.1002/rob.21541}.
\bibitem[{Redmon et~al.(2016)Redmon, Divvala, Girshick \& Farhadi}]{Redmon2016}
\bibinfo{author}{Redmon, J.}, \bibinfo{author}{Divvala, S.},
  \bibinfo{author}{Girshick, R.}, \& \bibinfo{author}{Farhadi, A.}
  (\bibinfo{year}{2016}).
\newblock \bibinfo{title}{You only look once: Unified, real-time object
  detection}.
\newblock In {\it \bibinfo{booktitle}{Proceedings of the IEEE conference on
  computer vision and pattern recognition}\/} (pp. \bibinfo{pages}{779--788}).
\bibitem[{Redmon \& Farhadi(2017)}]{Redmon2017}
\bibinfo{author}{Redmon, J.}, \& \bibinfo{author}{Farhadi, A.}
  (\bibinfo{year}{2017}).
\newblock \bibinfo{title}{Yolo9000: better, faster, stronger}.
\newblock In {\it \bibinfo{booktitle}{Proceedings of the IEEE conference on
  computer vision and pattern recognition}\/} (pp.
  \bibinfo{pages}{7263--7271}).
\bibitem[{Redmon \& Farhadi(2018)}]{Redmon2018}
\bibinfo{author}{Redmon, J.}, \& \bibinfo{author}{Farhadi, A.}
  (\bibinfo{year}{2018}).
\newblock \bibinfo{title}{{YOLO}~v3: {An Incremental Improvement [DB]}}.
\newblock {\it \bibinfo{journal}{arXiv preprint arXiv:1612.08242}\/}, .
\bibitem[{Ren et~al.(2015)Ren, He, Girshick \& Sun}]{NIPS2015_5638}
\bibinfo{author}{Ren, S.}, \bibinfo{author}{He, K.}, \bibinfo{author}{Girshick,
  R.}, \& \bibinfo{author}{Sun, J.} (\bibinfo{year}{2015}).
\newblock \bibinfo{title}{{Faster R-CNN: Towards Real-Time Object Detection
  with Region Proposal Networks}}.
\newblock In \bibinfo{editor}{C.~Cortes}, \bibinfo{editor}{N.~D. Lawrence},
  \bibinfo{editor}{D.~D. Lee}, \bibinfo{editor}{M.~Sugiyama}, \&
  \bibinfo{editor}{R.~Garnett} (Eds.), {\it \bibinfo{booktitle}{Advances in
  Neural Information Processing Systems 28}\/} (pp. \bibinfo{pages}{91--99}).
\newblock \bibinfo{publisher}{Curran Associates, Inc.}
\bibitem[{Rose et~al.(2016)Rose, Kicherer, Wieland, Klingbeil, T{\"{o}}pfer \&
  Kuhlmann}]{Rose2016}
\bibinfo{author}{Rose, J.}, \bibinfo{author}{Kicherer, A.},
  \bibinfo{author}{Wieland, M.}, \bibinfo{author}{Klingbeil, L.},
  \bibinfo{author}{T{\"{o}}pfer, R.}, \& \bibinfo{author}{Kuhlmann, H.}
  (\bibinfo{year}{2016}).
\newblock \bibinfo{title}{{Towards Automated Large-Scale 3D Phenotyping of
  Vineyards under Field Conditions}}.
\newblock {\it \bibinfo{journal}{Sensors}\/},  {\it \bibinfo{volume}{16}\/},
  \bibinfo{pages}{2136}. \DOIprefix\doi{10.3390/s16122136}.
\bibitem[{Roser(2019)}]{Roser}
\bibinfo{author}{Roser, M.} (\bibinfo{year}{2019}).
\newblock \bibinfo{title}{Employment in agriculture}.
\newblock {\it \bibinfo{journal}{Our World in Data}\/}, .
\newblock \bibinfo{note}{Https://ourworldindata.org/employment-in-agriculture}.
\bibitem[{Sa et~al.(2016)Sa, Ge, Dayoub, Upcroft, Perez \& McCool}]{Sa2016}
\bibinfo{author}{Sa, I.}, \bibinfo{author}{Ge, Z.}, \bibinfo{author}{Dayoub,
  F.}, \bibinfo{author}{Upcroft, B.}, \bibinfo{author}{Perez, T.}, \&
  \bibinfo{author}{McCool, C.} (\bibinfo{year}{2016}).
\newblock \bibinfo{title}{{DeepFruits: A Fruit Detection System Using Deep
  Neural Networks}}.
\newblock {\it \bibinfo{journal}{Sensors}\/},  {\it \bibinfo{volume}{16}\/},
  \bibinfo{pages}{1222}. \DOIprefix\doi{10.3390/s16081222}.
\bibitem[{Santos et~al.(2019)Santos, de~Souza, dos Santos~Andreza \&
  Avila}]{ZenodoWGISD}
\bibinfo{author}{Santos, T.}, \bibinfo{author}{de~Souza, L.},
  \bibinfo{author}{dos Santos~Andreza}, \& \bibinfo{author}{Avila, S.}
  (\bibinfo{year}{2019}).
\newblock \bibinfo{title}{[dataset] {Embrapa Wine Grape Instance Segmentation
  Dataset -- Embrapa WGISD}}.
\newblock \DOIprefix\doi{10.5281/zenodo.3361736}.
\bibitem[{Santos et~al.(2017)Santos, Bassoi, Oldoni \& Martins}]{Santos2017}
\bibinfo{author}{Santos, T.~T.}, \bibinfo{author}{Bassoi, L.~H.},
  \bibinfo{author}{Oldoni, H.}, \& \bibinfo{author}{Martins, R.~L.}
  (\bibinfo{year}{2017}).
\newblock \bibinfo{title}{{Automatic grape bunch detection in vineyards based
  on affordable 3D phenotyping using a consumer webcam}}.
\newblock In \bibinfo{editor}{J.~G.~A. Barbedo}, \bibinfo{editor}{M.~F. Moura},
  \bibinfo{editor}{L.~A.~S. Romani}, \bibinfo{editor}{T.~T. Santos}, \&
  \bibinfo{editor}{D.~P. Drucker} (Eds.), {\it \bibinfo{booktitle}{Anais do XI
  Congresso Brasileiro de Agroinform{\'{a}}tica (SBIAgro 2017)}\/} (pp.
  \bibinfo{pages}{89--98}).
\newblock \bibinfo{address}{Campinas}: \bibinfo{publisher}{Unicamp}.
\newblock \URLprefix
  \url{http://ainfo.cnptia.embrapa.br/digital/bitstream/item/169609/1/Automatic-grape-SBIAgro.pdf}.
\bibitem[{Satyanarayanan(2017)}]{Satyanarayanan2017}
\bibinfo{author}{Satyanarayanan, M.} (\bibinfo{year}{2017}).
\newblock \bibinfo{title}{The emergence of edge computing}.
\newblock {\it \bibinfo{journal}{Computer}\/},  {\it \bibinfo{volume}{50}\/},
  \bibinfo{pages}{30--39}.
\bibitem[{Scaramuzza \& Fraundorfer(2011)}]{Scaramuzza2011}
\bibinfo{author}{Scaramuzza, D.}, \& \bibinfo{author}{Fraundorfer, F.}
  (\bibinfo{year}{2011}).
\newblock \bibinfo{title}{{Visual Odometry [Tutorial]}}.
\newblock {\it \bibinfo{journal}{IEEE Robotics {\&} Automation Magazine}\/},
  {\it \bibinfo{volume}{18}\/}, \bibinfo{pages}{80--92}.
  \DOIprefix\doi{10.1109/MRA.2011.943233}.
\bibitem[{Sch\"{o}nberger \& Frahm(2016)}]{Schoenberger2016}
\bibinfo{author}{Sch\"{o}nberger, J.~L.}, \& \bibinfo{author}{Frahm, J.-M.}
  (\bibinfo{year}{2016}).
\newblock \bibinfo{title}{Structure-from-motion revisited}.
\newblock In {\it \bibinfo{booktitle}{Conference on Computer Vision and Pattern
  Recognition (CVPR)}\/}.
\bibitem[{Shelhamer et~al.(2017)Shelhamer, Long \& Darrell}]{Shelhamer2017}
\bibinfo{author}{Shelhamer, E.}, \bibinfo{author}{Long, J.}, \&
  \bibinfo{author}{Darrell, T.} (\bibinfo{year}{2017}).
\newblock \bibinfo{title}{{Fully Convolutional Networks for Semantic
  Segmentation}}.
\newblock {\it \bibinfo{journal}{IEEE Transactions on Pattern Analysis and
  Machine Intelligence}\/},  {\it \bibinfo{volume}{39}\/},
  \bibinfo{pages}{640--651}. \DOIprefix\doi{10.1109/TPAMI.2016.2572683}.
\bibitem[{Simonyan \& Zisserman(2015)}]{Simonyan2014}
\bibinfo{author}{Simonyan, K.}, \& \bibinfo{author}{Zisserman, A.}
  (\bibinfo{year}{2015}).
\newblock \bibinfo{title}{Very deep convolutional networks for large-scale
  image recognition}.
\newblock In {\it \bibinfo{booktitle}{{ICLR}}\/}.
\newblock \URLprefix \url{http://arxiv.org/abs/1409.1556}.
\bibitem[{Triggs et~al.(2000)Triggs, McLauchlan, Hartley \&
  Fitzgibbon}]{Triggs2000}
\bibinfo{author}{Triggs, B.}, \bibinfo{author}{McLauchlan, P.~F.},
  \bibinfo{author}{Hartley, R.~I.}, \& \bibinfo{author}{Fitzgibbon, A.~W.}
  (\bibinfo{year}{2000}).
\newblock \bibinfo{title}{{Bundle Adjustment — A Modern Synthesis Vision
  Algorithms: Theory and Practice}}.
\newblock In \bibinfo{editor}{B.~Triggs}, \bibinfo{editor}{A.~Zisserman}, \&
  \bibinfo{editor}{R.~Szeliski} (Eds.), {\it \bibinfo{booktitle}{Vision
  Algorithms: Theory and Practice}\/} \bibinfo{type}{book part (with own
  title)}~\bibinfo{chapter}{21}. (pp. \bibinfo{pages}{153--177}).
\newblock \bibinfo{publisher}{Springer Berlin / Heidelberg} volume
  \bibinfo{volume}{1883} of {\it \bibinfo{series}{Lecture Notes in Computer
  Science}\/}.
\newblock \DOIprefix\doi{10.1007/3-540-44480-7_21}.
\bibitem[{Van Der~Walt et~al.(2011)Van Der~Walt, Colbert \& Varoquaux}]{NumPy}
\bibinfo{author}{Van Der~Walt, S.}, \bibinfo{author}{Colbert, S.~C.}, \&
  \bibinfo{author}{Varoquaux, G.} (\bibinfo{year}{2011}).
\newblock \bibinfo{title}{The {NumPy} array: a structure for efficient
  numerical computation}.
\newblock {\it \bibinfo{journal}{Computing in Science \& Engineering}\/},  {\it
  \bibinfo{volume}{13}\/}, \bibinfo{pages}{22}.
\bibitem[{Vincent \& Soille(1991)}]{vincent1991watersheds}
\bibinfo{author}{Vincent, L.}, \& \bibinfo{author}{Soille, P.}
  (\bibinfo{year}{1991}).
\newblock \bibinfo{title}{Watersheds in digital spaces: an efficient algorithm
  based on immersion simulations}.
\newblock {\it \bibinfo{journal}{IEEE Transactions on Pattern Analysis \&
  Machine Intelligence}\/},  (pp. \bibinfo{pages}{583--598}).

\end{thebibliography}

\appendix

\section{Embrapa Wine Grape Instance Segmentation Dataset -- Embrapa
  WGISD}
\label{appendix:WGISD}

This section presents a detailed description of the dataset, a \emph{datasheet for the dataset} 
as proposed by \citet{Gebru2018}.

\subsection{Motivation for Dataset Creation}
\label{sec:org576a247}

\subsubsection{Why was the dataset created?}
\label{sec:org6e480f9}

Embrapa WGISD (\emph{Wine Grape Instance Segmentation Dataset}) was created
to provide images and annotation to study \emph{object detection and
instance segmentation} for image-based monitoring and field robotics
in viticulture. It provides instances from five different grape
varieties taken from the field. These instances show variance in grape
pose, illumination and focus, including genetic and phenological
variations such as shape, color and compactness. 

\subsubsection{What (other) tasks could the dataset be used for?}
\label{sec:orgfe05c1f}

Possible uses include relaxations of the instance segmentation
problem: classification (Is a grape in the image?), semantic
segmentation (What are the ``grape pixels'' in the image?), and object
detection (Where are the grapes in the image?). The WGISD can also be used
in grape variety identification.


\subsubsection{Who funded the creation of the dataset?}
\label{sec:org7c9950e}

The building of the WGISD dataset was supported by the Embrapa SEG
Project 01.14.09.001.05.04, \emph{Image-based metrology for Precision
Agriculture and Phenotyping}, and the CNPq PIBIC Program (grants 161165/2017-6 and 
125044/2018-6).

\subsection{Dataset Composition}
\label{sec:org9163256}

\subsubsection{What are the instances?}
\label{sec:org2c40949}

Each instance consists of an RGB image  and an annotation describing
grape clusters locations as bounding boxes. A subset of the instances
also contains binary masks identifying the pixels belonging to each
grape cluster. Each image presents at least one grape cluster. Some grape
clusters can appear far at the background and should be ignored. 

\subsubsection{Are relationships between instances made explicit in the data?}
\label{sec:org30a40eb}

File names prefixes identify the variety observed in the instance
(Table~\ref{table:GenInfoData}). 


\subsubsection{How many instances of each type are there?}
\label{sec:orgd3b5757}

The dataset consists of 300 images containing 4,432 grape clusters
identified by bounding boxes. A subset of 137 images also contains
binary masks identifying the pixels of each cluster. It means that
from the 4,432 clusters, 2,020 of them present binary masks for
instance segmentation, as summarized in
Table~\ref{table:GenInfoData}. 

\subsubsection{What data does each instance consist of?}
\label{sec:org3a7677d}

Each instance contains an 8-bit RGB image and a text file
containing one bounding box description per line. These text files
follow the ``YOLO format'' \citep{Redmon2016}:

\begin{verbatim}
CLASS CX CY W H
\end{verbatim}

\emph{class} is an integer defining the object class -- the dataset presents
only the grape class that is numbered 0, so every line starts with 
this ``class zero'' indicator. The center of the bounding box is the point
$(c_x, c_y)$, represented as float values because this format normalizes
the coordinates by the image dimensions. To get the absolute position,
use  $(2048 \cdot c_x, 1365 \cdot c_y)$. The bounding box dimensions are given by
$W$ and $H$, also normalized by the image size. 

The instances presenting mask data for instance segmentation contain
files presenting the \texttt{.npz} extension. These files are compressed
archives for NumPy $n$-dimensional arrays \citep{NumPy}. Each array is a 
$H \times W \times n_\mathrm{clusters}$ three-dimensional array
where $n_\mathrm{clusters}$  is the number of grape clusters observed in the image. After assigning the NumPy array to a variable \texttt{M}, the mask for the $i$-th grape cluster can be found in
\texttt{M[:,:,i]}. The $i$-th mask corresponds to the $i$-th line in the bounding
boxes file.

The dataset also includes the original image files, presenting
the full original resolution. The normalized annotation
for bounding boxes allows easy identification of clusters in the
original images, but the mask data will need to be properly 
rescaled if users wish to work on the original full resolution. 

\subsubsection{Is everything included or does the data rely on external resources?}
\label{sec:org9af0525}

Everything is included in the dataset.

\subsubsection{Are there recommended data splits or evaluation measures?}
\label{sec:org4be082b}

The dataset comes with specified train/test splits. The splits are
found in lists stored as text files. There are also lists referring
only to instances presenting binary masks. 


Standard measures from the information retrieval and computer vision
literature should be employed: precision and recall, $F_1$ score and
average precision as seen in \citet{MSCOCO} and \citet{PascalVOC}.

\subsubsection{What experiments were initially run on this dataset?}
\label{sec:org4b2d134}

To the present date, this work describes the first experiments run on
this dataset.

\subsection{Data Collection Process}
\label{sec:orgb6d3e5e}

\subsubsection{How was the data collected?}

Images were captured at the vineyards of Guaspari Winery, located at
Espírito Santo do Pinhal, São Paulo, Brazil (Lat -22.181018, Lon
-46.741618). The winery staff performs dual pruning: one for shaping
(after previous year harvest) and one for production, resulting in
canopies of lower density.  The image capture was realized in April
2017 for \emph{Syrah} and in April 2018 for the other varieties (see
Table~\ref{table:GenInfoData}).  No pruning, defoliation or any
intervention in the plants was performed specifically for the dataset
construction: the images capture a real, trellis system-based wine
grape production. The camera captures the images in a frontal pose,
that means the camera principal axis is approximately perpendicular to
the wires of the trellis system and the plants rows.

A Canon\texttrademark \,  EOS REBEL T3i DSLR camera and a
Motorola\texttrademark \, Z2 Play smartphone were used to 
capture the images. The cameras were located between the vines lines,
facing the vines at distances around 1-2~meters. The EOS REBEL T3i
camera captured 240 images, including all \emph{Syrah} pictures. The
Z2 smartphone grabbed 60 images covering all varieties except
\emph{Syrah}. The REBEL images were scaled to $2048 \times 1365$ pixels and
the Z2 images to $2048 \times 1536$ pixels (see Section~\ref{sec:WGISDPreProc}).
More data about the capture process can be found in the
Exif data found in the original image files, included in the dataset. 

\subsubsection{Who was involved in the data collection process?}

The authors of this paper. T.~T.~Santos, A.~A.~Santos and
S.~Avila captured the images in field. T.~T.~Santos, L.~L.~de~Souza 
and S.~Avila performed the annotation.

\subsubsection{How was the data associated with each instance acquired?}

The rectangular bounding boxes identifying the grape clusters were
annotated using the \texttt{labelImg}
tool\footnote{\url{https://github.com/tzutalin/labelImg}}. The clusters
can be under severe occlusion by leaves, trunks or other
clusters. Considering the absence of 3-D data and on-site annotation,
the clusters locations had to be defined using only a single-view image,
so some clusters could be incorrectly delimited.

A subset of the bounding boxes was selected for mask annotation, using
a novel tool developed by the authors and presented in this work. This
interactive tool lets the annotator mark grape and background pixels
using scribbles, and a graph matching algorithm developed by
\citet{Noma2012} is employed to perform image segmentation to every
pixel in the bounding box, producing a binary mask representing
grape/background classification.

\subsection{Data Preprocessing}

\subsubsection{What preprocessing/cleaning was done?}
\label{sec:WGISDPreProc}

The following steps were taken to process the data:

\begin{enumerate}
  \item Bounding boxes were annotated for each image using the
    \texttt{labelImg} tool.
  \item Images were resized to $W = 2048$ pixels. This
    resolution proved to be practical for mask annotation, a
    convenient balance between grape detail and time spent by the
    graph-based segmentation algorithm.
  \item A randomly selected subset of images was employed for mask
    annotation using the interactive tool based on graph matching.
  \item All binaries masks were inspected, in search of pixels
    attributed to more than one grape cluster. The annotator assigned
    the disputed pixels to the most likely cluster.
  \item The bounding boxes were fitted to the masks, which provided a
    fine-tuning of grape cluster locations. 
\end{enumerate}

\subsubsection{Was the ``raw'' data saved in addition to the
  preprocessed data?}

The original resolution images, containing the Exif data provided by
the cameras, is available in the dataset.

\subsection{Dataset Distribution}

\subsubsection{How is the dataset distributed?}

The dataset is available at Zenodo \citep{ZenodoWGISD}.

\subsubsection{When will the dataset be released/first distributed?}

The dataset was released in July 2019.

\subsubsection{What license (if any) is it distributed under?}

The data is released under Creative Commons BY-NC 4.0
(Attribution-NonCommercial 4.0 International license). There is a
request to cite the corresponding paper if the dataset is used. For
commercial use, contact the Embrapa Agricultural Informatics business
office at \texttt{cnptia.parcerias@embrapa.br}.

\subsubsection{Are there any fees or access/export restrictions?}

There are no fees or restrictions. For commercial use, contact Embrapa Agricultural 
Informatics business office at \texttt{cnptia.parcerias@embrapa.br}.

\subsection{Dataset Maintenance}

\subsubsection{Who is supporting/hosting/maintaining the dataset?}

The dataset is hosted at Embrapa Agricultural Informatics and all
comments or requests can be sent to Thiago T. Santos at
\texttt{thiago.santos@embrapa.br} (maintainer).

\subsubsection{Will the dataset be updated?}

There are no scheduled updates. In case of further updates, releases will be
properly tagged at GitHub.


\subsubsection{If others want to extend/augment/build on this dataset,
  is there a mechanism for them to do so?}

Contributors should contact the maintainer by e-mail.

\subsubsection{No warranty}

The maintainers and their institutions are \emph{exempt from any liability, judicial or extrajudicial, for any losses or damages arising from the use of the data contained in the image database}.

\end{document}